\documentclass[lettersize,journal]{IEEEtran}
\usepackage{amsmath,amsfonts}
\usepackage{algorithmic}
\usepackage{algorithm}
\usepackage{array}
\usepackage[caption=false,font=normalsize,labelfont=sf,textfont=sf]{subfig}
\usepackage{textcomp}
\usepackage{stfloats}
\usepackage{url}
\usepackage{verbatim}
\usepackage{graphicx}
\usepackage{cite}
\usepackage{multirow}
\usepackage[table,xcdraw]{xcolor}
\usepackage[normalem]{ulem}
\useunder{\uline}{\ul}{}
\usepackage{booktabs}
\usepackage{xspace}
\newcommand{\ours}[0]{\texttt{Grad2Fair}\xspace}
\newcommand{\oursm}[0]{\texttt{GradDist}\xspace}
\usepackage{ulem}
\newcommand{\code}[0]{\url{https://github.com/ZzoomD/Grad2Fair}}
\usepackage{amssymb}

\begin{document}

\title{Grad2Fair: A Gradient-driven Approach for Graph Fairness without Demographics}

\author{Yuchang Zhu, Zezhong Xie, Huizhe Zhang, Huazhen Zhong, Jintang Li, Liang Chen, and Zibin Zheng,~\IEEEmembership{Fellow,~IEEE,} 
\thanks{Manuscript received April 19, 2021; revised August 16, 2021.}
\thanks{The research is supported by the National Key R\&D Program of China under grant No. 2022YFF0902500, the Guangdong Basic and Applied Basic Research Foundation, China (No. 2023A1515011050), Shenzhen Science and Technology Program (KJZD20231023094501003), GMCC-SYSU Joint Lab for Smart Applications, and Tencent AI Lab (RBFR2024004). ({\itshape Corresponding author: Liang Chen.})}
\thanks{Liang Chen is with the School of Computer Science and Engineering, Sun Yat-Sen University, Guangzhou 510007, China. Email: chenliang6@mail.sysu.edu.cn}
\thanks{Zibin Zheng is with the School of Software Engineering, Sun Yat-sen University, Zhuhai  519082, China.}
\thanks{Jintang Li is with the Institute of Artificial Intelligence, Xiamen University, Xiamen 361005, China.}
\thanks{Yuchang Zhu, Zezhong Xie, Huizhe Zhang, and Huazhen Zhong are with the School of Computer Science and Engineering, Sun Yat-sen University, Guangzhou 510007, China.}}

\markboth{Journal of \LaTeX\ Class Files,~Vol.~14, No.~8, August~2021}%
{Shell \MakeLowercase{\textit{et al.}}: A Sample Article Using IEEEtran.cls for IEEE Journals}

\IEEEpubid{0000--0000/00\$00.00~\copyright~2021 IEEE}

\maketitle

\begin{abstract}
Graph neural networks (GNNs) frequently encounter group fairness issues, often yielding biased predictions against specific demographic groups defined by sensitive attributes such as gender or race. While this challenge has motivated extensive research, most existing solutions rely on the strong assumption that demographics are fully available. To bypass this strict requirement, a few recent studies have attempted to use predicted demographics as proxies to enforce fairness constraints. However, predicted demographics may be inaccurate, resulting in the failure to improve fairness. In this work, we investigate the problem of graph fairness without demographic information and avoid the utilization of predicted demographics. Motivated by our observation that the gradient distributions of misclassified nodes implicitly encode demographic information, we first propose \oursm, a gradient-based metric that quantifies bias by measuring the distance between local modes within these distributions. To mitigate this bias, we propose Gradient-to-Fairness (\ours), a gradient-guided approach for group fairness without demographics. Due to the potential demographics in gradients, \ours directly leverages gradients to debias and eliminates demographic prediction, thereby enabling stable fairness performance. Experiments on several real-world datasets demonstrate the effectiveness of \ours, as evidenced by superior performance over baselines in most cases. Our code is available at \code.
\end{abstract}


\section{Introduction}
\label{sec:introduction}
\IEEEPARstart{G}{raph}-structured data, consisting of graph topology and node features, is ubiquitous in real-world scenarios, including social networks~\cite{social}, traffic networks~\cite{zhao2019t}, and molecular structures~\cite{molecular}. While graph neural networks (GNNs)~\cite{gcn,graphsage,maskgae} have emerged as a powerful approach for modeling such data, recent studies~\cite{fairgnn,nifty,fairdrop} reveal a critical concern: GNNs yield biased predictions against demographic groups defined by sensitive attributes, e.g., gender and race. This phenomenon is referred to as group unfairness and poses significant ethical risks, hindering the deployment of GNNs in high-stakes applications.

\begin{figure}[t]
  \centering
  \begin{picture}(450,145)
  \put(0,0){\includegraphics[width=\linewidth]{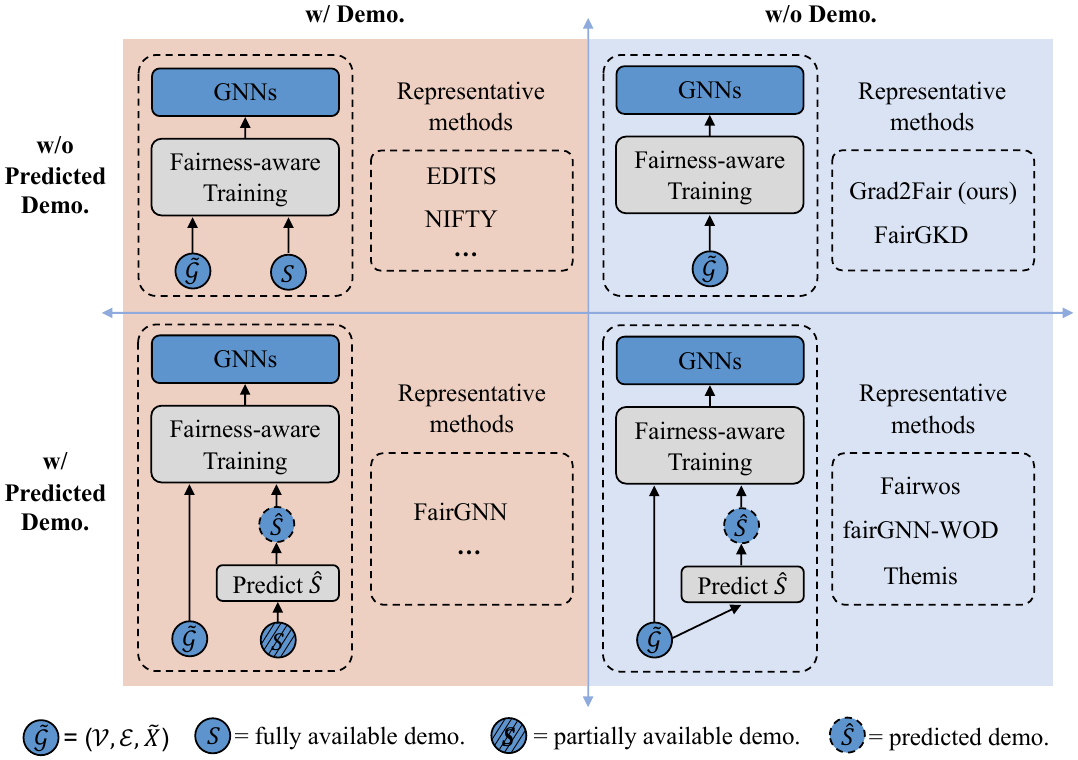}}
  \put(116,135.2){\tiny {\cite{edits}}}
  \put(116,125){\tiny {\cite{nifty}}}
  \put(118,56.5){\tiny {\cite{fairgnn}}}
  \put(225.5,121.4){\tiny {\cite{fairgkd}}}
  \put(223.5,62.6){\tiny {\cite{fairwos}}}
  \put(232.5,52.5){\tiny {\cite{fairgnnwod}}}
  \put(223,41.7){\tiny {\cite{themis}}}
  \end{picture}
  \caption{A taxonomy of representative graph fairness methods, categorized by demographic availability and the utilization of inferred demographics. ``Demo.'' is an abbreviation for demographics. ``w/'' and ``w/o'' denote ``with'' and ``without'', respectively. The pink region indicates areas outside the scope of this paper.}
  \label{fig:repre_methods}
\end{figure}
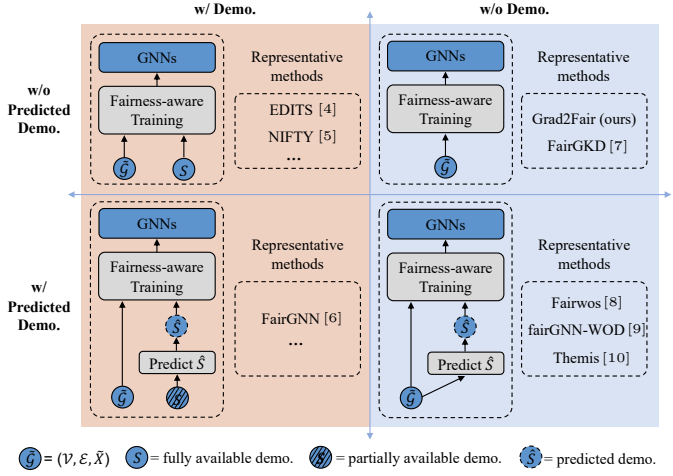

To alleviate this phenomenon, a number of studies have been conducted to improve the fairness of trained GNNs. A popular paradigm is adversarial training~\cite{fairgnn,fairvgnn,graphair}, which seeks to learn fair representations by ensuring that node embeddings are invariant to sensitive attributes through minimax optimization. In addition to adversarial approaches, diverse strategies have been employed to mitigate bias, including group distribution distance minimization~\cite{edits}, disentanglement~\cite{fairsad}, re-balancing~\cite{fairgb}, neutralization~\cite{fairsin}, and alignment~\cite{fairdla}. In summary, the core idea behind these methods is to eliminate sensitive attribute information from model outcomes through well-designed fairness constraints. Despite their success, these methods rely heavily on the availability of sensitive attributes, as shown in the pink areas of Fig.~\ref{fig:repre_methods}. However, in real-world scenarios, sensitive attributes, also known as \textit{demographics}, may be unavailable due to stringent privacy regulations. For example, the General Data Protection Regulation (GDPR)~\cite{gdpr} strictly mandates the protection of personal data concerning racial origin, political opinions, or religious beliefs. Consequently, developing fairness-aware GNNs in the absence of demographics has emerged as a critical and urgent challenge for real-world applications.

\IEEEpubidadjcol

To bridge this gap, recent studies~\cite{fairgnnwod,themis} attempt to achieve fairness without demographics. As shown in the blue areas of Fig.~\ref{fig:repre_methods}, existing methods can be categorized into two streams based on whether they use predicted demographic information. The first and more popular stream, e.g., Fairwos~\cite{fairwos}, follows a predict-then-constrain pipeline, which first predicts demographics and then enforces fairness constraints based on this predicted information. However, the inevitable discrepancy between inferred demographics and the ground truth often renders such constraints ineffective or counterproductive. To avoid this limitation, a second line of work, including FairGKD~\cite{fairgkd} and our proposed method, seeks to mitigate bias without explicit demographic inference. While FairGKD~\cite{fairgkd} explores the relationship between biases and partial data training, e.g., using only node features, it primarily targets high-level bias mitigation. As a result, addressing fairness-related biases is often treated as a byproduct rather than a primary objective, ultimately leading to marginal improvements in fairness.

\begin{figure}[t]
  \centering
  \includegraphics[width=\linewidth]{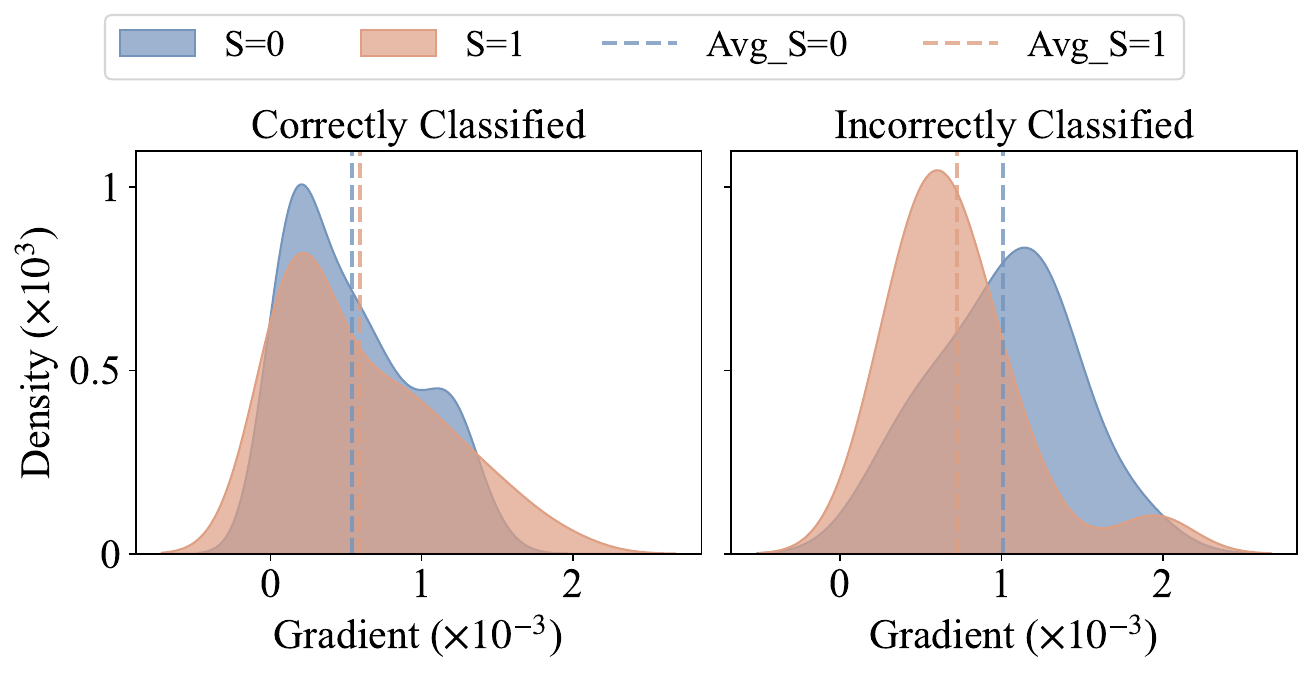}
  \caption{Kernel Density Estimation (KDE) of gradient distributions for correctly and incorrectly classified samples. While gradients for correctly classified instances overlap, those within the misclassified subset exhibit pronounced distributional divergence across different demographic groups.}
  \label{fig:preliminary}
\end{figure}

A recent study~\cite{gog} suggests that gradients are more effective in representing sensitive attributes, a phenomenon which has been primarily validated on tabular data but remains under-explored in the graph domain. Motivated by this, we investigate the gradient behavior of misclassified samples in graph-structured data. As shown in Fig.~\ref{fig:preliminary}, we conduct a statistical analysis of gradient distributions and observe that different demographic groups exhibit distinct gradient distributions within the subset of misclassified nodes. Remarkably, in some cases, these gradient profiles possess sufficient discriminative power to allow for the accurate inference of demographic labels. Further details regarding this empirical observation are provided in Section~\ref{sec:emp_metric}. Consequently, exploiting the demographic signals naturally embedded within gradients offers a principled approach to enhancing algorithmic fairness.

In this work, we explore achieving fairness through gradients in the absence of demographic information. First, we introduce \oursm, a bias evaluation metric that quantifies the degree of unfairness by measuring the distributional divergence between peaks in the gradient density distributions. Second, we propose a simple yet effective approach named \uline{Grad}ient-\uline{to}-\uline{Fair}ness (\ours) to ensure fairness without demographics. Specifically, \ours consists of two stages: bias amplification and upweighting via gradient. The first stage intentionally leverages shortcut learning to amplify inherent model biases. Building upon these amplified signals, the second stage mitigates bias through upweighting training based on gradients. Through this design, \ours maintains a minimal architectural footprint, offering a high-performance yet computationally efficient solution for fair GNNs. Our contributions are as follows:
\begin{itemize}
    \item We introduce \oursm, an empirically driven metric for bias evaluation in the absence of demographics. Based on the insights from Fig.~\ref{fig:preliminary}, \oursm quantifies algorithmic bias by measuring the inter-modal distance between the peaks of gradient density distributions.
    \item We propose \ours, a novel approach for fairness without demographics. Inspired by the discovered correlation between gradient distributions and demographics, \ours adaptively reweights the training objective to prioritize informative yet challenging samples. 
    \item We first provide an empirical validation of \oursm, followed by a systematic evaluation of \ours from both theoretical and empirical standpoints. Experimental results demonstrate that \ours outperforms state-of-the-art baselines in most cases, while maintaining superior computational efficiency.
\end{itemize}

The remainder of this paper is organized as follows. Section~\ref{sec:rela_work} provides a brief overview of related literature. Section~\ref{sec:prelimi} introduces the necessary notations and formalizes the problem definition. Section~\ref{sec:emp_metric} presents an empirical investigation and introduces a bias evaluation metric for fairness without demographics. Section~\ref{sec:method} introduces details of our proposed approach \ours, followed by the complexity analysis. Section~\ref{sec:exp} presents experimental results to verify the effectiveness of \ours. Finally, Section~\ref{sec:conclusion} summarizes this paper and highlights its limitations.

\section{Related Work}
\label{sec:rela_work}
In this section, we provide a brief overview of related literature, including group fairness in graph learning and fairness without demographic information.

\subsection{Group Fairness in Graph Learning}
Group fairness methods in graph learning aim to develop fair graph algorithms that provide equitable outcomes across groups defined by sensitive attributes, such as race or age. While early research focused on traditional graph mining~\cite{fairwalk}, the emergence of GNNs~\cite{hgnn,graphsage} has shifted the focus toward ensuring fairness within deep graph models. These studies can be categorized into pre-processing and in-processing methods. Pre-processing methods focus on mitigating biases within the data itself to provide clean inputs for training. For example, FairDrop~\cite{fairdrop} utilizes a biased edge dropout algorithm to reduce homophily with respect to the sensitive attribute. EDITS~\cite{edits} minimizes the Wasserstein distance between groups to debias node attributes and topology. Similarly, Graphair~\cite{graphair} employs adversarial learning to guide data debiasing, and FairAGG~\cite{fairagg} reweights edges based on their fairness contributions via Shapley values. In contrast, in-processing methods integrate fairness constraints directly into the training process. FairGNN~\cite{fairgnn} and FairVGNN~\cite{fairvgnn} leverage adversarial learning to learn representations that are independent of sensitive attributes. NIFTY~\cite{nifty} incorporates counterfactual contrastive learning to ensure model invariance, while FairGB~\cite{fairgb} employs counterfactual node mixup and contribution alignment loss to rebalance groups during training. FairINV~\cite{fairinv} reframes group fairness as invariant learning and employs an invariant objective. To address the potential utility degradation caused by fairness-aware training, FairSAD~\cite{fairsad}, FUGNN~\cite{fugnn}, and FairSIN~\cite{fairsin} investigate the trade-off between utility and fairness. Beyond these core areas, research has also expanded into fairness under distribution shifts~\cite{FatraGNN}, fair graph transformers~\cite{fairgt,fairgp}, fair federated graph learning~\cite{fedGL}, and fairness considering false positives~\cite{fairgse}. 

Despite significant progress, most existing pre-processing and in-processing methods rely on the strong assumption that demographic information is fully available. In practice, however, such information is frequently inaccessible due to stringent privacy regulations and legal constraints. To address this, \ours aims to achieve fairness without demographic information, thereby providing a solution for this practical scenario.

\subsection{Fairness without Demographics}
Driven by strict privacy regulations, fairness without demographics~\cite{SPECTRE} has emerged as a pivotal research frontier in machine learning. One prominent research trajectory exploits the relationship between non-sensitive features and sensitive attributes. Recognizing that features highly correlated with sensitive attributes act as proxies, several studies, e.g., FairRF~\cite{fairrf}, minimize the correlation between non-sensitive features and final predictions to achieve fairness. Meanwhile, another line of studies focuses on Rawlsian Max-Min fairness~\cite{Rawlsianfairness} without demographics, which maximizes the performance of the worst-off groups. For example, distributionally robust optimization (DRO)~\cite{dro} employs a robust optimization framework to enhance performance for high-loss subpopulations. To address the susceptibility of DRO to noise, adversarial reweighted learning (ARL)~\cite{arl} identifies and improves computationally identifiable groups via adversarial training. Additionally, recent advancements~\cite{zhao2022fundamental} have further integrated paradigms like invariant representation learning and knowledge distillation to improve Rawlsian Max-Min fairness without demographics. However, these methods have been predominantly validated on tabular data, leaving their effectiveness on graph-structured data under-explored.

In graphs, research on fairness without demographics can be classified into two paradigms, depending on whether predicted demographics are used. The first paradigm follows a two-stage approach, i.e., initially predicting demographics and subsequently enforcing fairness constraints based on these surrogates. Representative methods, including Fairwos~\cite{fairwos}, fairGNN-WOD~\cite{fairgnnwod}, and Themis~\cite{themis}, fall into this paradigm. However, these methods inevitably suffer from inaccurate demographic predictions, which limits their effectiveness in improving fairness. The second paradigm bypasses explicit demographic predictions by learning fair GNNs directly. For example, building upon the insight that partial data training can mitigate bias, FairGKD~\cite{fairgkd} leverages a knowledge distillation framework, guiding a student model via a fair teacher trained on partial data. Nevertheless, FairGKD~\cite{fairgkd} treats group fairness as a collateral benefit of high-level bias mitigation, which frequently results in suboptimal fairness gains. In contrast, \ours belongs to this latter paradigm but introduces a more principled approach. By delving into the relationship between gradients and unknown sensitive attributes, \ours performs bias mitigation tailored for fairness-related bias, significantly outperforming FairGKD in terms of fairness performance. 

\section{Preliminaries}
\label{sec:prelimi}

\subsection{Notations}
For clarity, we provide a brief introduction to the notations related to our work. We focus on the node classification task and denote $Y\in \{0,1\}^n$ as the node label vector. Let $\mathcal{G} = (\mathcal{V}, \mathcal{E}, X)$ denote an undirected attributed graph, where $\mathcal{V}$ is a set of $|\mathcal{V}| = n$ nodes and $\mathcal{E}$ is a set of $|\mathcal{E}| = m$ edges. $X\in \mathbb{R}^{n \times d}$ represents the node feature matrix with $d$ dimensions. The adjacency matrix $A\in \{0,1\}^{n\times n}$ describes the graph topology, where $A_{uv}=1$ if an edge $e_{uv}\in \mathcal{E}$ connects node $u$ and $v$, and $A_{uv}=0$ otherwise. $S\in\{0,1\}^{n}$ represents the binary sensitive attribute, where $S_{u}=S_{v}$ indicates that nodes $u$ and $v$ belong to the same demographic group. $\Tilde{X}\in \mathbb{R}^{n \times (d-1)}$ is the node feature matrix without $S$.

\subsection{Group Fairness in GNNs}
Group fairness in GNNs focuses on the equity of model predictions across demographic groups. For clarity, we take binary classification as an example to describe group fairness in GNNs. Given an undirected attributed graph $\mathcal{G} = (\mathcal{V}, \mathcal{E}, X)$, a GNN classifier $f(\cdot)$ takes $\mathcal{G}$ as input and predicts node labels $\hat{Y}\in \{0,1\}^{n}$. The group fairness can be evaluated using demographic parity (DP)~\cite{DP} and equal opportunity (EO)~\cite{EO}. Specifically, demographic parity requires predictions to be independent of the sensitive attribute $S$, and equal opportunity ensures the same true positive rate for each demographic group. Group fairness of GNNs is commonly measured by the DP and EO differences between two demographic groups, which can be defined as follows: 
\begin{equation}
    \label{eq:metric}
    \resizebox{0.9\linewidth}{!}{%
    $\begin{aligned}
    \Delta_{DP} &= \lvert P(\hat{Y}_v=1 \mid S_v=0) - P(\hat{Y}_v=1 \mid S_v=1) \rvert, \\
    \Delta_{EO} &= \lvert P(\hat{Y}_v=1 \mid Y_v=1, S_v=0) - P(\hat{Y}_v=1 \mid Y_v=1, S_v=1) \rvert,
    \end{aligned}$%
    }
    \end{equation}
where small $\Delta_{DP}$ and $\Delta_{EO}$ indicate fairer GNN models.

\subsection{Problem Definition}
In this work, we investigate the problem of graph fairness without demographics, which requires trained models to make fair predictions without access to demographic information during the training, validation, and testing phases. Specifically, given an undirected attributed graph $\Tilde{\mathcal{G}}=(\mathcal{V}, \mathcal{E}, \Tilde{X})$ without the sensitive attribute $S$, along with the node label ground truth $Y$, our goal is to learn a fair GNN classifier $f(\cdot)$ on $\Tilde{\mathcal{G}}$. Thus, the optimization problem for group fairness without demographics can be defined as follows:
\begin{equation}
    \begin{aligned}
        \label{definition}
        \hat{Y} &= f_{\theta}(\Tilde{\mathcal{G}}), \\
        \theta^{*} &= \arg\min_{\theta}\ \mathcal{L}_{\mathrm{cls}}\!\big(\hat{Y}, Y\big) \quad \text{s.t.} \quad \mathcal{C}_{\mathrm{fair}}\!\big(\hat{Y}; \Tilde{\mathcal{G}}\big) \le \epsilon,
    \end{aligned}
\end{equation}
where $\theta$ is the trainable parameter of the model $f(\cdot)$ and $\theta^{*}$ is the optimal parameter. $\mathcal{L}_{\mathrm{cls}}(\cdot,\cdot)$ represents the loss function for the node classification task, and $\mathcal{C}_{\mathrm{fair}}(\cdot;\cdot)$ represents a fairness constraint function. $\epsilon$ represents a fairness tolerance threshold, indicating the maximum tolerable fairness disparity.

\section{Empirical Investigation and Our Proposed Metric}
\label{sec:emp_metric}
Existing methods to improve group fairness without demographics first predict demographic information, which is employed to enforce fairness constraints. However, these predicted demographic attributes often deviate from the true demographic distribution, resulting in suboptimal fairness improvements or even fairness degradation. To bridge this gap, a natural approach is to avoid the utilization of predicted demographics. Consequently, in this section, we conduct a preliminary study to obtain more insights into fairness without demographics. Then, we introduce a metric named \oursm for the bias evaluation without demographics.

\subsection{Empirical Investigation}
\label{emp}
A recent study~\cite{gog} reveals that gradients are effective cues for representing demographic information. While this phenomenon has been verified on tabular data, it remains under-explored on graph-structured data. Motivated by this, we conduct a gradient investigation to analyze the correlation between gradients and demographics. Our investigations can be summarized into two parts, i.e., experiments on synthetic and real-world datasets. We first conduct a qualitative experiment on synthetic datasets with varying bias levels. Subsequently, we further verify our observations through an experiment on real-world datasets. 

\begin{figure*}[!t]
  \centering
  \includegraphics[width=\linewidth]{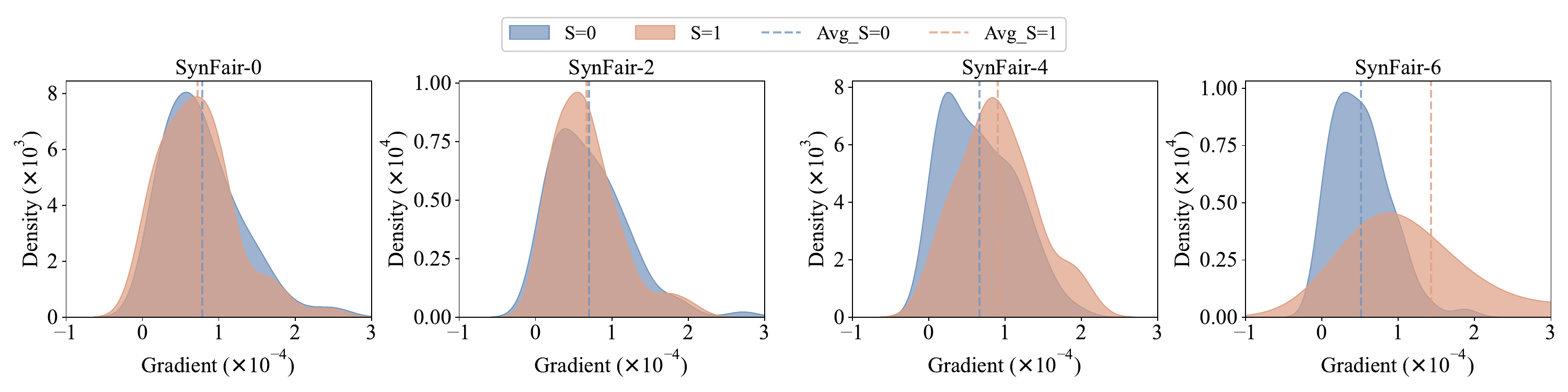}
  \caption{Gradient visualization of misclassified nodes on synthetic datasets with different bias levels. The gradient average between demographic groups exhibits a larger margin on a more biased dataset, e.g., SynFair-6.}
  \label{fig:preli1_grad}
\end{figure*}

\begin{table}[!t]
\caption{Correlations between misclassified node gradients and bias levels.}
\label{preli_corr}
\centering
\renewcommand\arraystretch{1.1}
\begin{tabular}{cccc}
\toprule
                                         & \textbf{Pearson (r)} & \textbf{Spearman ($\rho$)} & \textbf{Kendall ($\tau$)} \\
                                         \midrule
$\Delta_{|p_{00}-p_{01}|}-\Delta_{grad}$ & 0.9014  & 0.8571   & 0.7143  \\
$\Delta_{DP}-\Delta_{grad}$              & 0.9301  & 0.9286   & 0.8095  \\
\bottomrule
\end{tabular}
\end{table}

For the experiment on synthetic datasets, we follow the settings in~\cite{fairness_benchmark} to generate synthetic datasets with different bias levels. Specifically, these synthetic datasets have binary sensitive attributes and node labels. We set the probability $p_{00}$ of $S=0, Y=0$ to vary from 0.25 to 0.07 with an interval of 0.03, where $p_{11}=p_{00}$, $p_{01}=p_{10}$, $p_{11}+p_{00}+p_{01}+p_{10}=1$. Meanwhile, we fix other parameters to align the settings of Syn-1 in~\cite{fairness_benchmark}. For clarity, the seven generated synthetic datasets are referred to as ``SynFair0'' to ``SynFair6''. According to Syn-2 in~\cite{fairness_benchmark}, the larger unbalanced group ratio, i.e., larger $|p_{00}-p_{01}|$, indicates a more significant unfair prediction and a higher bias level. Consequently, the unfairness degree of ``SynFair0/1/2/3/4/5/6'' varies from low to high. Based on these synthetic datasets, we train a 2-layer graph convolution network (GCN) classifier and then calculate differences $\Delta_{grad}$ of average gradient between two demographic groups within misclassified nodes on the training set. As shown in Table~\ref{preli_corr}, we report three types of correlation results: the correlation between the unbalanced group ratio $\Delta_{|p_{00}-p_{01}|}$ and $\Delta_{grad}$, and the correlation between the fairness performance $\Delta_{DP}$ and $\Delta_{grad}$. According to the definition in~\cite{1996Straightforward,1990Rank,0Statistical}, the experimental results demonstrate that $\Delta_{grad}$ exhibits strong positive correlations with $\Delta_{|p_{00}-p_{01}|}$ and $\Delta_{DP}$, respectively. $\Delta_{|p_{00}-p_{01}|}$ and $\Delta_{DP}$ are similar metrics for indicating bias levels, with higher values indicating greater bias. Motivated by these strong correlation results, a natural question arises: \textit{Do the gradient distributions of misclassified nodes within the training set reveal demographic information?} To this end, we visualize gradient distributions on various synthetic datasets, as shown in Fig.~\ref{fig:preli1_grad}. As the degree of unfairness of synthetic datasets increases, i.e., from SynFair0 to SynFair6, the gradient distribution between different demographic groups exhibits less overlap, indicating the potential demographic information within the gradient distribution.

\begin{figure}[!t]
  \centering
  \includegraphics[width=\linewidth]{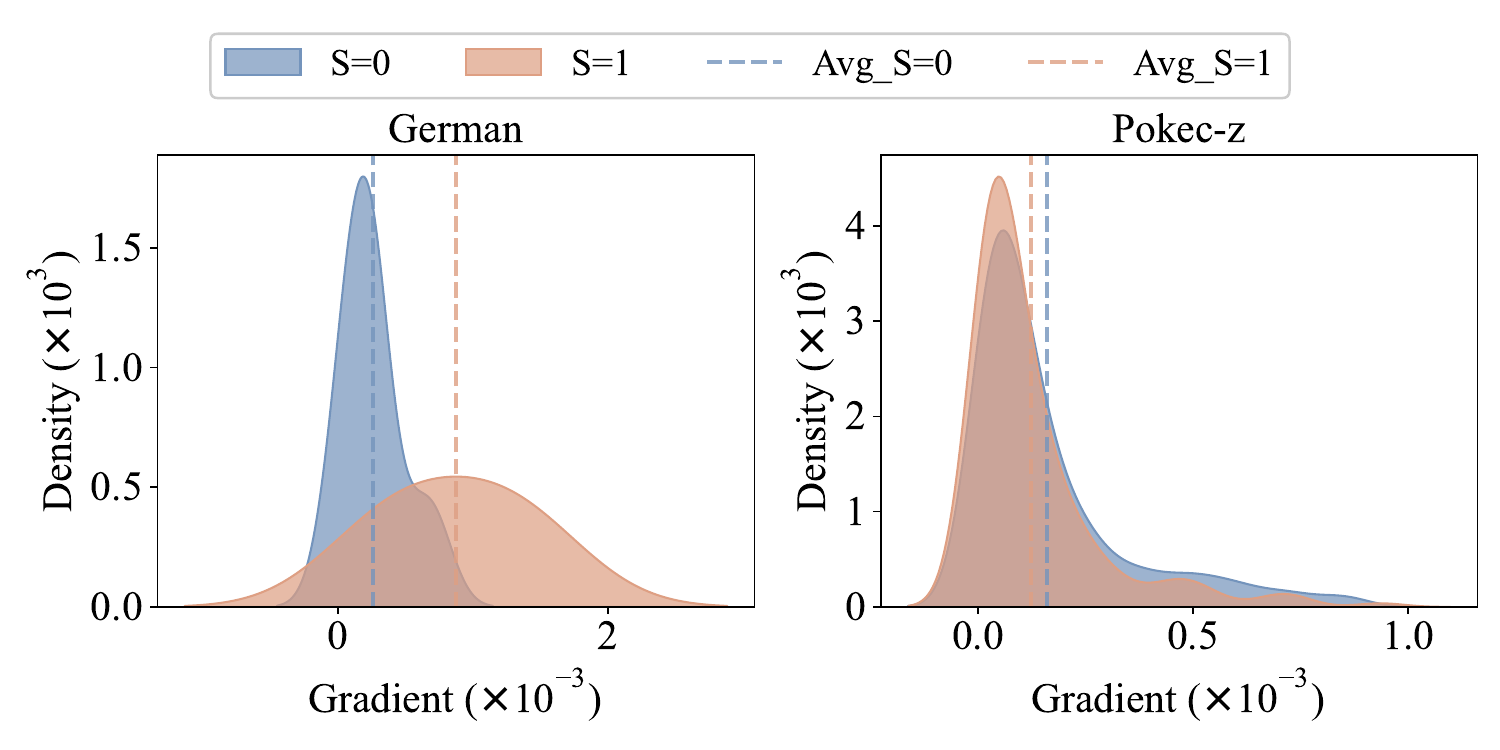}
  \caption{Gradient visualization of misclassified nodes on real-world datasets. The difference in average gradients between demographic groups varies significantly across datasets.}
  \label{fig:preli2_grad}
\end{figure}

For the experiment on real-world datasets, we visualize the gradient distribution of misclassified nodes on two commonly used real-world datasets, i.e., German and Pokec-Z datasets. As shown in Fig.~\ref{fig:preli2_grad}, we observe distinct gradient distributions on these two datasets. On the German dataset, the gradient distributions of different demographic groups exhibit little overlap. Conversely, a complete overlap occurs for the gradient distributions between two demographic groups on the Pokec-z dataset, indicating less demographic information.

In summary, we observe that gradient average differences between two demographic groups exhibit strong correlations with the degree of unfairness of datasets. Furthermore, the gradient distribution of misclassified nodes within the training set includes potential demographic information, paving the way to achieve fairness without demographic information.

\subsection{Bias Evaluation Metric}
\label{metric}
Inspired by our observation on the correlation between $\Delta_{grad}$ and bias levels, we introduce a bias evaluation metric for bias measurement without demographics, named \oursm. When demographic information is available, bias can be measured by calculating the average gradient difference $\Delta_{grad}$ between two demographic groups. However, calculating $\Delta_{grad}$ is infeasible when such information is unavailable. As demonstrated by the results on SynFair4 and 6 in Fig.~\ref{fig:preli1_grad}, gradient density curves for different demographic groups exhibit distinct peaks. Furthermore, the overlap between these peaks diminishes as the bias within the dataset increases. The core idea behind \oursm is that a biased model often exhibits distinct learning behaviors across different demographic subgroups, which manifest as distinct clusters (modes) in the gradient space. Thus, given the number of demographic groups, we can search for the position of peaks within gradient density curves and then calculate the difference between peaks. 

For simplicity, we take binary sensitive attributes as an example. Considering a set of observed discrete gradients $\mathcal{Q}_{g} = \{g_1, g_2, \dots, g_M\}$ extracted from a specific layer or loss function during training, we first seek to recover the underlying continuous probability density function (PDF) from these gradient observations. Since the true distribution is a priori unknown, we employ Kernel Density Estimation (KDE) to produce a non-parametric estimate. The estimated PDF $\hat{f}(g)$ is defined as:
\begin{equation}
\hat{f}(g) = \frac{1}{Mh} \sum_{i=1}^{M} K\left(\frac{g - g_i}{h}\right),
\end{equation}
where $M$ represents the number of nodes in the observed discrete gradient set. $K(\cdot)$ denotes the standard Gaussian kernel $K(u) = \frac{1}{\sqrt{2\pi}} e^{-\frac{1}{2}u^2}$ and $h > 0$ is the smoothing bandwidth. The infinite differentiability of the Gaussian kernel ensures that the resulting density curve is smooth, allowing for rigorous derivative-based analysis of the distribution's topography. In our implementation, we adopt a bandwidth of $h=0.15$ to strike an optimal balance between capturing local modalities and suppressing stochastic noise.

Upon obtaining the continuous PDF $\hat{f}(g)$, the next step involves characterizing the local modes that signify group-specific gradient concentrations. These peaks represent the most frequent gradient magnitudes, physically corresponding to the feature clustering centers of different latent populations. To identify these modes, we treat the peak-finding task as a constrained optimization problem. Specifically, a point $g^*$ is identified as a representative peak (local maximum) if it satisfies the following calculus-based criteria:
\begin{equation}
\frac{d}{dg} \hat{f}(g) \bigg|_{g=g^*} = 0 \quad \text{and} \quad \frac{d^2}{dg^2} \hat{f}(g) \bigg|_{g=g^*} < 0.
\end{equation}

The first-order condition (stationarity) ensures that the point is a critical point, where the summation $\sum_{i=1}^{M} K(\frac{g - g_i}{h}) \cdot (g - g_i)$ vanishes. The second-order condition (concavity) distinguishes local maxima from minima or saddle points, ensuring that the density curve is locally downward-opening. By solving these conditions, we obtain a set of candidate peaks $\mathcal{P} = \{g^*_j\}$. To capture the most significant bias-induced divergence between opposing subgroups, we extract the two most dominant peaks based on their density magnitudes:
\begin{equation}
\text{TopPeaks} = \{g^*_1, g^*_2\} \in \arg \text{top-2}_{g^* \in \mathcal{P}} \left( \hat{f}(g^*) \right),
\end{equation}

Building upon these identified peak coordinates, we define a distance-based metric to quantify the degree of polarization in a scale-invariant manner. We introduce \oursm, which measures the separation between the dominant modes relative to the total empirical range of the observed gradient:
\begin{equation}
\oursm = \frac{|g^*_1 - g^*_2|}{\max(\mathcal{Q}_{g}) - \min(\mathcal{Q}_{g})},
\end{equation}
where the denominator serves as a normalization factor that eliminates the influence of varying gradient magnitudes across different tasks or training stages, allowing for cross-model comparisons. A lower \oursm value indicates that the gradients of different samples are concentrated within a unified manifold, suggesting consistent optimization paths for different demographic groups. Conversely, a high \oursm indicates that the model updates are significantly disparate for different demographic groups. In summary, \oursm serves as a robust indicator for measuring algorithmic bias.

\section{Methodology}
\label{sec:method}
\begin{figure*}[!t]
  \centering
  \includegraphics[width=0.85\linewidth]{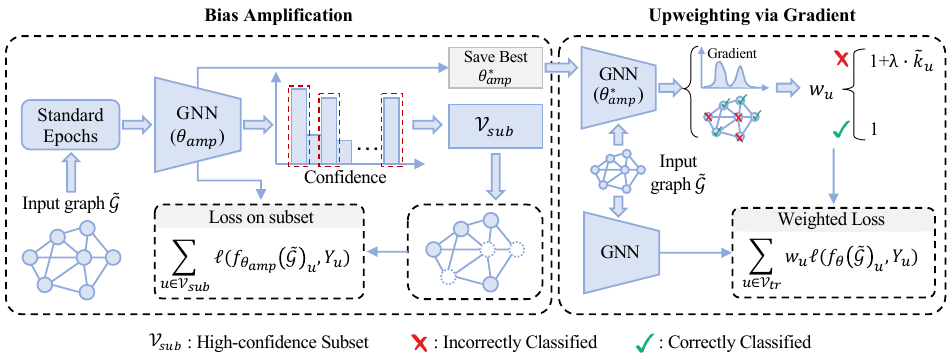}
  \caption{Framework overview of \ours.}
  \label{fig:overview}
\end{figure*}

In this section, we first propose \ours, followed by a theoretical analysis demonstrating how it improves fairness without demographics. Subsequently, we present the algorithm and complexity analysis to detail our proposed method. As shown in Fig.~\ref{fig:overview}, the core idea of \ours is to amplify bias and then debias by minimizing gradient disparity between demographic groups. Specifically, \ours consists of two stages: bias amplification and upweighting via gradients. The bias amplification stage leverages shortcut learning to guide the model to focus on high-confidence samples. This makes the trained model more biased towards the majority group, thereby amplifying bias. Building upon the gradients obtained from the bias amplification stage, the upweighting via the gradient stage upweights the loss contribution of misclassified samples, which steers the model's focus toward the minority group, thereby improving overall fairness.

\subsection{Bias Amplification}
\label{bias_amp}
As revealed by our empirical investigation in Section~\ref{emp}, the gradient distribution disparity ($\Delta_{grad}$) of misclassified nodes exhibits a strong positive correlation with both the dataset's inherent bias levels and the resulting fairness metrics, e.g., $\Delta_{DP}$. However, as illustrated in Fig. \ref{fig:preli2_grad}, certain real-world graph datasets, e.g., Pokec-z, exhibit significant overlap in the gradient distributions of different demographic groups. In such scenarios, the demographic information included within the gradient signals is insufficient, making it challenging for the model to effectively identify and mitigate latent biases through gradient information.

To address this challenge, we introduce a bias amplification strategy. Specifically, the core idea behind this strategy is that since the discriminative power of gradient disparity depends on the degree of bias, we can artificially amplify the initial biases captured by the model during the early stages of training. This strategy induces a more distinct gradient distribution that serves as a clear upweighting signal for subsequent debiasing.

As observed in prior studies~\cite{failure}, models initially fit bias-aligned samples before addressing bias-conflicting ones during the training stage. This process reflects the capture of data shortcuts. Consequently, models predict labels for bias-aligned samples with high confidence. We leverage this shortcut learning mechanism to develop a bias amplification strategy. Specifically, we select high-confidence samples to optimize the trained model during the initial training phase, thereby reinforcing the model’s reliance on these shortcuts. 

Let $f_{\theta_{amp}}$ denote a GNN classifier with trainable parameters $\theta_{amp}$. Given an undirected attributed graph $\Tilde{\mathcal{G}}=(\mathcal{V},\mathcal{E},\tilde{X})$, for each node $u \in \mathcal{V}_{tr}$, the prediction confidence is defined as $c_u = \max(f_{\theta_{amp}}( \Tilde{\mathcal{G}})_u)$. After a specific number of warm-up epochs, we select a subset of samples $\mathcal{V}_{sub}$ consisting of those with the highest confidence scores:
\begin{equation}
\mathcal{V}_{sub} = \{u \in \mathcal{V}_{tr} \mid c_u > \text{percentile}(c, 1-\tau) \},
\end{equation}
where $\tau$ is a hyperparameter representing the selection ratio, e.g., 0.25. The optimization objective for this stage is defined as:
\begin{equation}
\mathcal{L}_{amp} = \sum_{u \in \mathcal{V}_{sub}} \ell(f_{\theta_{amp}}(\tilde{\mathcal{G}})_u, Y_u),
\end{equation}
where $\ell(\cdot)$ denotes the cross-entropy loss. By iteratively optimizing $\mathcal{L}_{amp}$, the model is forced to overfit the shortcut patterns present in the high-confidence samples, thereby amplifying inherent bias.

\textbf{Theoretical Analysis of Bias Amplification.} We provide a theoretical justification for why optimizing $\mathcal{L}_{amp}$ on the high-confidence subset $\mathcal{V}_{sub}$ amplifies the inherent bias. Specifically, we model the training dynamics under the shortcut learning paradigm. Let the training nodes be partitioned into two underlying disjoint sets based on the dataset's latent bias: the bias-aligned (majority) group $\mathcal{V}_{A}$ and the bias-conflicting (minority) group $\mathcal{V}_{C}$. 

\textit{Assumption 1 (Confidence Disparity via Shortcut Learning).} Due to the shortcut mechanisms during the warm-up epochs, the GNN model fits the bias-aligned samples much faster than the bias-conflicting ones. Consequently, the confidence distribution is heavily skewed, satisfying:
\begin{equation}
P(u \in \mathcal{V}_{A} \mid c_u > \tau) \gg P(u \in \mathcal{V}_{C} \mid c_u > \tau).
\end{equation}

This implies that the selected subset $\mathcal{V}_{sub}$ is overwhelmingly dominated by $\mathcal{V}_{A}$, such that $|\mathcal{V}_{sub} \cap \mathcal{V}_{A}| \approx |\mathcal{V}_{sub}|$ and $|\mathcal{V}_{sub} \cap \mathcal{V}_{C}| \approx 0$. Based on Assumption 1, we can define the severity of the model's bias as the expected loss disparity between the bias-conflicting and bias-aligned groups, denoted as $\Delta_{loss}(\theta) = \mathbb{E}_{v \in \mathcal{V}_{C}}[\ell_v(\theta)] - \mathbb{E}_{u \in \mathcal{V}_{A}}[\ell_u(\theta)]$. The amplification of bias is equivalent to the strict increase of $\Delta_{loss}(\theta)$ during the optimization of $\mathcal{L}_{amp}$.

\textit{Theorem 1 (Bias Amplification via Skewed Gradient Flow).} Given a sufficiently small learning rate $\eta > 0$ and the optimization step $\theta_{t+1} = \theta_t - \eta \nabla_{\theta} \mathcal{L}_{amp}(\theta_t)$, optimizing the model on the high-confidence subset $\mathcal{V}_{sub}$ strictly increases the bias severity, i.e., $\Delta_{loss}(\theta_{t+1}) > \Delta_{loss}(\theta_t)$.

\textit{Proof Sketch.} Following Assumption 1, the gradient of the average amplification loss $\mathcal{L}_{amp}$ is dominated by the bias-aligned samples:
\begin{equation}
\nabla_{\theta} \mathcal{L}_{amp}(\theta_t) \approx \frac{1}{|\mathcal{V}_{sub}|} \sum_{u \in \mathcal{V}_{sub} \cap \mathcal{V}_{A}} \nabla_{\theta} \ell_u(\theta_t).
\end{equation}

Assuming the loss function $\ell$ is $L$-smooth, applying the Taylor expansion to the loss of an arbitrary node $x$ at step $t+1$ gives:
\begin{equation}
\ell_x(\theta_{t+1}) = \ell_x(\theta_t) - \eta \langle \nabla_{\theta} \ell_x(\theta_t), \nabla_{\theta} \mathcal{L}_{amp}(\theta_t) \rangle + R_x,
\end{equation}
where the remainder $R_x$ is uniformly bounded by $|R_x| \le \frac{L}{2} \eta^2 \|\nabla_{\theta} \mathcal{L}_{amp}(\theta_t)\|^2$.

Because the optimization direction is dominated by $\mathcal{V}_{A}$, we assume there exist constants $\alpha > 0$ and $\beta \ge 0$ with $\alpha > \beta$ such that the expected gradient correlations satisfy:
\begin{equation}
\mathbb{E}_{u \in \mathcal{V}_{A}}[\langle \nabla_{\theta} \ell_u(\theta_t), \nabla_{\theta} \mathcal{L}_{amp}(\theta_t) \rangle] \ge \alpha \|\nabla_{\theta} \mathcal{L}_{amp}(\theta_t)\|^2,
\end{equation}
\begin{equation}
\mathbb{E}_{v \in \mathcal{V}_{C}}[\langle \nabla_{\theta} \ell_v(\theta_t), \nabla_{\theta} \mathcal{L}_{amp}(\theta_t) \rangle] \le \beta \|\nabla_{\theta} \mathcal{L}_{amp}(\theta_t)\|^2.
\end{equation}

Taking the expectation of the loss difference for both demographic groups, we obtain:
\begin{align}
    \begin{split}
    \mathbb{E}_{u \in \mathcal{V}_{A}}[\ell_u(\theta_{t+1}) - \ell_u(\theta_t)] \le &-\eta \alpha \|\nabla_{\theta} \mathcal{L}_{amp}(\theta_t)\|^2 \\
    &+ \frac{L}{2} \eta^2 \|\nabla_{\theta} \mathcal{L}_{amp}(\theta_t)\|^2,
    \end{split}
\end{align}
\begin{align}
    \begin{split}
    \mathbb{E}_{v \in \mathcal{V}_{C}}[\ell_v(\theta_{t+1}) - \ell_v(\theta_t)] \ge &-\eta \beta \|\nabla_{\theta} \mathcal{L}_{amp}(\theta_t)\|^2 \\
    & - \frac{L}{2} \eta^2 \|\nabla_{\theta} \mathcal{L}_{amp}(\theta_t)\|^2.
    \end{split}
\end{align}

Subtracting the expected change of the bias-aligned group from that of the bias-conflicting group yields the change in bias severity:
\begin{align}
    \begin{split}
    \Delta_{loss}(\theta_{t+1}) - \Delta_{loss}(\theta_t) \ge &\eta (\alpha - \beta) \|\nabla_{\theta} \mathcal{L}_{amp}(\theta_t)\|^2\\
    &- L \eta^2 \|\nabla_{\theta} \mathcal{L}_{amp}(\theta_t)\|^2.
    \end{split}
\end{align}

For a sufficiently small learning rate satisfying $\eta < \frac{\alpha - \beta}{L}$, the first-order inner product terms strictly dominate the remainder bounds. Consequently, we obtain:
\begin{equation}
\Delta_{loss}(\theta_{t+1}) - \Delta_{loss}(\theta_t) > 0
\end{equation}
which directly implies $\Delta_{loss}(\theta_{t+1}) > \Delta_{loss}(\theta_t)$. In summary, the theoretical analysis above proves the effectiveness of our bias amplification strategy, which provides a foundation for the subsequent debiasing. $\hfill \blacksquare$

\subsection{Upweighting via Gradient}
\label{upweighting}
The bias amplification strategy described in Section~\ref{bias_amp} successfully bridges the gap between latent demographic groups and observable gradient signals. Specifically, under the bias-amplified model $f_{\theta_{amp}}$, different demographic groups exhibit distinct gradient distributions, even without explicit access to sensitive attributes $S$. This enables the gradient space to serve as an effective proxy for identifying latent minority groups, providing a novel pathway for achieving fairness without relying on demographics. Unlike existing methods that rely on latent variable discovery via auxiliary models, e.g., VAEs, which are prone to error propagation, our approach utilizes the amplified gradient sensitivity as a sample-specific weighting signal.

Given an undirected attributed graph $\Tilde{\mathcal{G}}=(\mathcal{V}, \mathcal{E}, \Tilde{X})$, the bias-amplified model $f_{\theta_{amp}}$ quantifies the fitting difficulty and feature sensitivity of each node. To derive a node-specific importance score, we compute the gradient of the loss function with respect to the input features $\tilde{X}$ for each node $u \in \mathcal{V}_{tr}$:
\begin{equation}
g_u = \nabla_{\tilde{X}_u} \ell(f_{\theta_{amp}}(\Tilde{\mathcal{G}})_u, Y_u), \quad k_u = \|g_u\|_2.
\end{equation}

We adopt the gradient norm relative to input features rather than model parameters because the former directly captures the local geometry of the decision boundary around each specific sample. As shown in Theorem 1, since the bias amplification stage forces the decision boundary to overfit the majority group's shortcut patterns, nodes from the minority group (bias-conflicting samples) are pushed into regions with high loss and sharp gradient transitions, resulting in larger $k_u$ values.

To balance the contributions across groups while maintaining the model's overall utility, we focus on the set of misclassified nodes $\mathcal{M} = \{u \in \mathcal{V}_{tr} \mid \text{argmax}(f_{\theta_{amp}}(\Tilde{\mathcal{G}})_u) \neq Y_u\}$, as these nodes are most likely to be victims of model bias. We apply min-max normalization to the gradient norms within $\mathcal{M}$ to ensure stability across different datasets:
\begin{equation}
\tilde{k}_u = \frac{k_u - \min_{v \in \mathcal{M}}(k_v)}{\max_{v \in \mathcal{M}}(k_v) - \min_{v \in \mathcal{M}}(k_v) + \epsilon},
\end{equation}
where $\epsilon$ is a small constant for numerical stability. We then assign an importance weight $w_u$ to each node $u$:
\begin{equation}
\label{eq:upweight}
w_u = 
\begin{cases} 
1 + \lambda \cdot \tilde{k}_u, & \text{if } u \in \mathcal{M}, \\
1, & \text{otherwise},
\end{cases}
\end{equation}
where $\lambda > 0$ is a hyperparameter controlling the strength of the debiasing. Finally, the model is trained by minimizing the weighted objective function:
\begin{equation}
\mathcal{L}_{upweight} = \sum_{u \in \mathcal{V}_{tr}} w_u \cdot \ell(f_{\theta}(\tilde{\mathcal{G}})_u, Y_u).
\end{equation}

By assigning higher weights to misclassified nodes with larger gradient norms, the model is guided to refine its decision boundary for the minority group, thereby improving fairness.

\textbf{Theoretical Analysis of Upweighting via Gradient.} We extend our theoretical framework in Section~\ref{bias_amp} to formally demonstrate how minimizing $\mathcal{L}_{upweight}$ mitigates the bias amplified in the previous stage. Recall that $\mathcal{V}_{A}$ and $\mathcal{V}_{C}$ denote the bias-aligned (majority) and bias-conflicting (minority) groups, respectively, and bias severity is defined as $\Delta_{loss}(\theta) = \mathbb{E}_{v \in \mathcal{V}_{C}}[\ell_v(\theta)] - \mathbb{E}_{u \in \mathcal{V}_{A}}[\ell_u(\theta)]$.

\textit{Assumption 2 (Weight Disparity via Amplified Gradients).} Due to the bias amplification stage (Theorem 1), nodes in $\mathcal{V}_{C}$ are predominantly misclassified and exhibit sharper gradient transitions, i.e., larger $k_u$. Consequently, the assigned importance weights for the minority group are strictly larger in expectation than those for the majority group:
\begin{equation}
\mathbb{E}_{v \in \mathcal{V}_{C}}[w_v] \gg \mathbb{E}_{u \in \mathcal{V}_{A}}[w_u].
\end{equation}

\textit{Assumption 3 (Gradient Alignment via Upweighting).} While Assumption 2 guarantees weight disparity, we further assume that the weighted global gradient flow aligns more closely with the descent direction of the heavily weighted minority group. Formally, we assume there exist constants $\gamma_C > \gamma_A > 0$ such that the expected inner products between individual sample gradients and the overall upweighted gradient $\nabla_{\theta} \mathcal{L}_{upweight}(\theta)$ satisfy:
\begin{equation}
\small
\mathbb{E}_{v \in \mathcal{V}_{C}}[\langle \nabla_{\theta} \ell_v(\theta), \nabla_{\theta} \mathcal{L}_{upweight}(\theta) \rangle] \ge \gamma_C \|\nabla_{\theta} \mathcal{L}_{upweight}(\theta)\|^2,
\end{equation}
\begin{equation}
\small
\mathbb{E}_{u \in \mathcal{V}_{A}}[\langle \nabla_{\theta} \ell_u(\theta), \nabla_{\theta} \mathcal{L}_{upweight}(\theta) \rangle] \le \gamma_A \|\nabla_{\theta} \mathcal{L}_{upweight}(\theta)\|^2.
\end{equation}

This assumption captures the typical optimization dynamics in upweighting strategies: upweighting a specific subset naturally forces the overall gradient update to be strongly correlated with the average gradient of that subset.

\textit{Theorem 2 (Bias Mitigation via Weighted Gradient Flow).} Under Assumptions 2 and 3, given a sufficiently small learning rate $\eta > 0$, and the optimization step $\theta_{t+1} = \theta_t - \eta \nabla_{\theta} \mathcal{L}_{upweight}(\theta_t)$, optimizing the weighted objective strictly decreases the bias severity, i.e., $\Delta_{loss}(\theta_{t+1}) < \Delta_{loss}(\theta_t)$.

\textit{Proof Sketch.} The gradient of the upweighting objective is a weighted sum of individual gradients. Under Assumption 3, the optimization direction $\nabla_{\theta} \mathcal{L}_{upweight}(\theta_t)$ is highly correlated with the descent direction of $\mathcal{V}_{C}$ ($\gamma_C > \gamma_A$). 

Assuming the loss function $\ell$ is $L$-smooth, applying the Taylor expansion to an arbitrary node $x$ yields $\ell_x(\theta_{t+1}) = \ell_x(\theta_t) - \eta \langle \nabla_{\theta} \ell_x(\theta_t), \nabla_{\theta} \mathcal{L}_{upweight}(\theta_t) \rangle + R_x$, with the remainder uniformly bounded by $|R_x| \le \frac{L}{2} \eta^2 \|\nabla_{\theta} \mathcal{L}_{upweight}(\theta_t)\|^2$.

Taking the expectation of the loss changes for both groups, we establish the upper bound for the minority group and the lower bound for the majority group based on the correlation bounds defined in Assumption 3:
\begin{align}
    \begin{split}
    \mathbb{E}_{v \in \mathcal{V}_{C}}[\ell_v(\theta_{t+1}) - \ell_v(\theta_t)] \le &-\eta \gamma_C \|\nabla_{\theta} \mathcal{L}_{upweight}(\theta_t)\|^2 \\
    &+ \frac{L}{2} \eta^2 \|\nabla_{\theta} \mathcal{L}_{upweight}(\theta_t)\|^2,
    \end{split}
\end{align}
\begin{align}
    \begin{split}
    \mathbb{E}_{u \in \mathcal{V}_{A}}[\ell_u(\theta_{t+1}) - \ell_u(\theta_t)] \ge &-\eta \gamma_A \|\nabla_{\theta} \mathcal{L}_{upweight}(\theta_t)\|^2 \\
    &- \frac{L}{2} \eta^2 \|\nabla_{\theta} \mathcal{L}_{upweight}(\theta_t)\|^2.
    \end{split}
\end{align}

Subtracting the expected change of the majority group from that of the minority group bounds the change in bias severity:
\begin{align}
    \begin{split}
    \Delta_{loss}(\theta_{t+1}) - \Delta_{loss}(\theta_t) \le &-\eta (\gamma_C - \gamma_A) \|\nabla_{\theta} \mathcal{L}_{upweight}(\theta_t)\|^2 \\
    &+ L \eta^2 \|\nabla_{\theta} \mathcal{L}_{upweight}(\theta_t)\|^2.
    \end{split}
\end{align}

For a sufficiently small learning rate satisfying $\eta < \frac{\gamma_C - \gamma_A}{L}$, the first-order negative term dominates, yielding:
\begin{equation}
\Delta_{loss}(\theta_{t+1}) - \Delta_{loss}(\theta_t) < 0.
\end{equation}

This process establishes that optimizing under the proposed upweighting strategy steadily reduces the expected loss disparity between the minority and majority groups, formally validating its effectiveness in mitigating the amplified bias. $\hfill \blacksquare$

\subsection{Algorithm and Complexity Analysis}
In this subsection, we summarize the training algorithm of our proposed method and then provide a theoretical analysis of its efficiency in terms of time and space complexity.

\subsubsection{Algorithm} Algorithm~\ref{alg:overall} presents a detailed training process of \ours. Specifically, the training process consists of two stages: 1) \textit{Bias Amplification}: an initial phase to amplify bias by optimizing over a high-confidence subset. 2) \textit{Upweighting via Gradient}: a subsequent debiasing phase via a weighted loss function derived from the gradient information of the bias-amplified model. 

\begin{algorithm}[t]
\caption{Training Algorithm of \ours}
\label{alg:overall}
\begin{algorithmic}[1]
\STATE \textbf{Input:} Undirected attributed graph $\tilde{\mathcal{G}} = (\mathcal{V}, \mathcal{E}, \tilde{X})$, labels $Y$, hyperparameters $\tau$ (selection ratio), $\lambda$ (debiasing strength), $\epsilon$ (stability constant), bias amplification epochs $E_{amp}$, upweighting epochs $E_{upweight}$.
\STATE \textbf{Output:} Fairness-enhanced model parameters $\theta^*$.
\STATE 
\STATE \textit{// Stage 1: Bias Amplification}
\STATE Initialize parameters $\theta_{amp}$ for the bias-amplified model.
\FOR{$epoch = 1$ \TO $E_{amp}$}
    \STATE Compute prediction confidence $c_u = \max(f_{\theta_{amp}}(\tilde{\mathcal{G}})_u)$ for all $u \in \mathcal{V}_{tr}$.
    \STATE Identify high-confidence subset: $\mathcal{V}_{sub} = \{u \in \mathcal{V}_{tr} \mid c_u > \text{percentile}(c, 1-\tau) \}$.
    \STATE Update $\theta_{amp}$ by minimizing $\mathcal{L}_{amp} = \sum_{u \in \mathcal{V}_{sub}} \ell(f_{\theta_{amp}}(\tilde{\mathcal{G}})_u, Y_u)$.
\ENDFOR
\STATE Save the best amplified model parameters $\theta_{amp}^*$.
\STATE 
\STATE \textit{// Stage 2: Upweighting via Gradient}
\STATE Load $\theta_{amp}^*$ and identify the misclassified set $\mathcal{M} = \{u \in \mathcal{V}_{tr} \mid \text{argmax}(f_{\theta_{amp}^*}(\Tilde{\mathcal{G}})_u) \neq Y_u\}$.
\STATE Compute input gradients $g_u = \nabla_{\tilde{X}_u} \ell(f_{\theta_{amp}^*}(\tilde{\mathcal{G}})_u, Y_u)$ for $u \in \mathcal{M}$.
\STATE Calculate gradient norms $k_u = \|g_u\|_2$ and apply min-max normalization to obtain $\tilde{k}_u$ using $\epsilon$.
\STATE Generate importance weights $W = \{w_u\}_{u \in \mathcal{V}_{tr}}$ where $w_u = 1 + \lambda \cdot \tilde{k}_u$ if $u \in \mathcal{M}$, and $w_u = 1$ otherwise.
\STATE Initialize parameters $\theta$ for the final fairness-enhanced model.
\FOR{$epoch = 1$ \TO $E_{upweight}$}
    \STATE Update $\theta$ by minimizing the weighted objective $\mathcal{L}_{upweight} = \sum_{u \in \mathcal{V}_{tr}} w_u \cdot \ell(f_{\theta}(\tilde{\mathcal{G}})_u, Y_u)$.
\ENDFOR
\RETURN Optimized parameters $\theta^* = \theta$.
\end{algorithmic}
\end{algorithm}

\subsubsection{Complexity Analysis}
\label{sec:complexity}
Let $n = |\mathcal{V}|$, $m = |\mathcal{E}|$, $L$ be the number of GNN layers, and $d$ be the hidden dimension. Here, we analyze the complexity of \ours.

\textit{Time Complexity:} The time complexity of \ours is primarily determined by the GNN forward and backward passes, taking $\mathcal{O}(L(md + nd^2))$ per epoch. Meanwhile, the bias amplification stage requires $E_{amp}$ epochs, where the per-epoch sorting cost $\mathcal{O}(n \log n)$ for subset selection is asymptotically dominated and negligible. The upweighting via gradient stage requires one additional forward and backward pass to compute input gradients for $\mathcal{M}$, followed by $E_{upweight}$ training epochs. The weight assignment has a complexity of $\mathcal{O}(nd)$, which is negligible. Overall, the total time complexity is $\mathcal{O}((E_{amp} + E_{upweight} + 1) \cdot L(md + nd^2))$. The single extra pass is a one-time cost. Thus, \ours introduces no asymptotic time overhead compared to standard GNN training.

\textit{Space Complexity:} The baseline memory footprint includes the graph structure $\mathcal{O}(m + nd)$ and model parameters $\mathcal{O}(Ld^2)$. During the upweighting via gradient stage, calculating input gradients requires temporary memory, but we only persistently cache the derived scalar importance weights $w_u$, reducing the extra overhead to just $\mathcal{O}(n)$. Consequently, the overall space complexity remains $\mathcal{O}(m + nd + Ld^2)$. \ours introduces only a marginal constant-factor memory increase, preserving the space complexity of the base architecture.

\section{Experiments}
\label{sec:exp}

In this section, we evaluate our method on the node classification task using four widely used real-world datasets and three synthetic datasets. We compare \ours with three state-of-the-art baselines for fair node classification, namely FairGKD~\cite{fairgkd}, Fairwos~\cite{fairwos}, and FDKD~\cite{fdkd}, across two representative GNN backbones.

\subsection{Experimental Setup}

\textit{Datasets:}
We conduct experiments on four widely used real-world datasets and three synthetic datasets. Real-world datasets include Bail~\cite{Bail}, Credit~\cite{Credit}, Pokec-z~\cite{Pokec}, and Pokec-n~\cite{Pokec}, which cover diverse application scenarios. For all real-world datasets, we adopt a standard train/validation/test split ratio of 50\%/25\%/25\% for node partitioning. For synthetic datasets, we generate SynFair following the setting of~\cite{fairness_benchmark}, while AttrBias and StruBias are constructed based on the protocols in~\cite{edits}. We split these synthetic datasets into train/validation/test sets according to the ratio of 60\%/20\%/20\%. Detailed descriptions of the real-world datasets are as follows:

\begin{itemize}
    \item \textbf{Bail}: A judicial decision-making dataset containing records of defendants released on bail during the period 1990–2009. Nodes correspond to these defendants, and edges are built according to the similarity of defendants' personal demographics and past criminal history records. The task is to classify whether defendants are on bail or not, with ``race" as the sensitive attribute.
    \item \textbf{Credit}: A real-world dataset focusing on credit card user payment behavior, providing detailed information about users' credit accounts and historical payment records. Nodes are credit card users, and edges are formed based on the similarity of users' payment behavior and account information. The task is to predict future credit card payment defaults, with ``age" as the sensitive attribute.
    \item \textbf{Pokec-z/Pokec-n}: Two subsets sampled from Pokec, the most popular social network in Slovakia, with anonymized user data collected in 2012. These two subsets are partitioned based on users' geographic provinces. Nodes represent social network users, and edges are precomputed from the inherent social connections between platform users. The task is to infer users' working fields, with ``region" as the sensitive attribute.
\end{itemize}

Detailed descriptions of the synthetic datasets are as follows:
\begin{itemize}
    \item \textbf{SynFair}: SynFair consists of seven subsets with different degrees of unfairness, referred to as ``SynFair0/1/2/3/4/5/6'', varying from a low degree of unfairness to a high degree. We generate SynFair following the setting of ``Syn-1'' in~\cite{fairness_benchmark}. For each subset, we set the number of nodes to 5,000. To generate subsets with different unfairness, we vary the probability $p_{00}$ (where $S=0$ and $Y=0$) from 0.25 to 0.07 with an interval of 0.03, where $p_{11}=p_{00}$, $p_{01}=p_{10}$, $p_{11}+p_{00}+p_{01}+p_{10}=1$. The generated datasets correspond to ``SynFair0/1/2/3/4/5/6''.
    \item \textbf{AttrBias/StruBias}: AttrBias and StruBias are synthetic datasets with different attribute bias and structural bias, respectively. We generate AttrBias and StruBias adopting the synthetic configurations for biased attributes and structures detailed in~\cite{edits}. Specifically, both datasets consist of 1,000 nodes, with the binary sensitive attribute evenly distributed. The binary labels for the downstream task are derived from the sum of the third and fourth feature dimensions, combined with injected Gaussian noise. This formulation ensures that the ground-truth targets are theoretically independent of the sensitive attribute.
    \begin{itemize}
        \item \textbf{AttrBias}: This dataset simulates a scenario with severe attribute bias but a completely fair graph structure. Each node is assigned a 10-dimensional feature vector. To inject attribute bias, the first two feature dimensions are drawn from distinct Gaussian distributions based on the sensitive group: $\mathcal{N}(-\mu, 1^2)$ for $S=0$ and $\mathcal{N}(\mu, 1^2)$ for $S=1$. To generate datasets with different bias levels, we vary $\mu$ in the range of $\{1,2,3,4,5,6,7\}$. The remaining eight dimensions are drawn from an unbiased uniform distribution $\mathcal{U}(0, 1)$. The network structure is generated using an Erdős-Rényi random graph model with a uniform edge probability of $p=0.002$, ensuring the topology remains agnostic to the demographic groups.
        \item \textbf{StruBias}: This dataset simulates a scenario characterized by extreme structural homophily and group isolation, while maintaining perfectly unbiased node features. Each node is assigned a 10-dimensional feature vector. The first two feature dimensions are independently drawn from the same standard Gaussian, and the rest of the features are sampled from uniform distributions. To construct the biased topology, we rank nodes based on the sum of their first two feature dimensions. We can control the intra-community edge probability $p_{intra}$ to generate StruBias. 
        We vary the intra-community edge probability $p_{intra}$ in the range of $\{0.05,0.1,0.15,0.2,0.25,0.3,0.35\}$, while setting the inter-community edge probability $p_{inter}$ and $k$ to 0.0001 and 250, resulting in seven datasets with different bias levels.
    \end{itemize}
\end{itemize}

\textit{Evaluation Metrics:} We evaluate the performance of \ours from two complementary perspectives, i.e., utility and fairness. For utility evaluation, we leverage F1-score and ACC as evaluation metrics, where higher values indicate better utility performance. For fairness evaluation, we use two classic fairness metrics, i.e., $\Delta_{\text{DP}}$~\cite{DP} and $\Delta_{\text{EO}}$~\cite{EO}, which can be computed as Eq.~\eqref{eq:metric}. Smaller $\Delta_{\text{DP}}$ and $\Delta_{\text{EO}}$ values indicate better group fairness.

\textit{Baselines:} We compare \ours with three state-of-the-art baseline methods for fair node classification, namely, FairGKD~\cite{fairgkd}, Fairwos~\cite{fairwos}, and FDKD~\cite{fdkd}.

\begin{itemize}
    \item \textbf{FairGKD}~\cite{fairgkd}: A demographic-agnostic method for learning fair GNNs without accessing sensitive attributes during training. Motivated by the observation that training on partial data (only node attributes or only graph topology) improves fairness, it constructs a synthetic teacher by combining fairness experts trained on partial data and uses knowledge distillation to guide the student GNN.
    \item \textbf{Fairwos}~\cite{fairwos}: A counterfactual fairness framework for GNNs that operates in the absence of explicit sensitive attributes. It first generates pseudo-sensitive attributes via an encoder to capture the latent influence of sensitive information, then finds realistic graph counterfactuals from the dataset to serve as fairness constraints. 
    \item \textbf{FDKD}~\cite{fdkd}: A fairness method that leverages knowledge distillation and soft label reweighting without requiring demographic information. It trains an overfitted teacher model and uses its normalized logits as soft labels to guide a student model. Theoretically, this acts as an error-based reweighting mechanism to focus on challenging yet correctly classified samples. Although FDKD is not specifically designed for graph data, we adapt it to the graph setting by replacing its backbone with a GNN.
\end{itemize}

\textit{Implementation Details:} 
We run all experiments five times and report the average results. For all methods, we utilize a 1-layer GCN or a 1-layer GIN as the backbone, followed by a linear layer as the classifier. The hidden dimensions for all backbones and the classifier are set to 16. We use the Adam optimizer with a weight decay of $1\times10^{-5}$ across all methods. All experiments are conducted on an NVIDIA GeForce RTX 3090 GPU, and all models are implemented with PyTorch and PyTorch-Geometric. We perform a grid search to obtain the optimal hyperparameters for all methods. The learning rate is searched from \{0.01, 0.001\} for all methods. Detailed settings are provided as follows: 
\begin{itemize}
    \item \textbf{\ours}: We perform a grid search for bias amplification epochs $E_{amp}$, upweighting epochs $E_{upweight}$, and debiasing strength $\lambda$ in a range of 100 to 700 with a step size of 50, \{500, 600\}, and 1 to 20 with a step size of 1, respectively. The selection ratio $\tau$ is set to 0.5. To ensure accurate shortcut capture by the bias-amplified model, we train it for 50 or 100 standard epochs during the initial phase of bias amplification. 
    \item \textbf{FairGKD}~\cite{fairgkd}: We conduct a grid search for the scalar temperature parameter $\tau$ and the disadvantaged loss enhancement parameter $\gamma$ in a range of \{0.001, 0.01, 0.1, 0.5, 1\}. The training epochs are set to 1000.
    \item \textbf{Fairwos}~\cite{fairwos}: We conduct a grid search for $\alpha$ and $K$ in a range of \{0.01, 0.05, 1, 2, 5\} and \{1, 2, 5, 10, 20\}, respectively. The training epochs for the first and second processes are set to 1000 and 15, respectively.
    \item \textbf{FDKD}~\cite{fdkd}: We perform a grid search for  trade-off hyperparameter $\alpha$ and temperature $T$ in a range of \{0.1, 0.3, 0.5, 0.7, 0.9\} and \{0.1, 0.3, 0.5, 0.7, 0.9\}, respectively. The training epochs of the teacher and the student model are set to 1500 and 1000, respectively.
\end{itemize}

\subsection{Overall Performance}
In this subsection, we first verify the effectiveness of \oursm and then compare \ours with several baselines over two commonly used GNN backbones.

\subsubsection{Effectiveness of \oursm}
\begin{table}[!t]
\caption{Correlations between results of \oursm and bias-controlling parameters.}
\label{graddist_effect}
\centering
\renewcommand\arraystretch{1.0}
\begin{tabular}{cccc}
\toprule
                                         & \textbf{Pearson (r)} & \textbf{Spearman ($\rho$)} & \textbf{Kendall ($\tau$)} \\
                                         \midrule
\textbf{SynFair}     & 0.8722  & 0.8214   & 0.6190  \\
\textbf{AttrBias}    & 0.8983  & 0.9643   & 0.9048  \\
\textbf{StruBias}  & 0.8917  & 0.8214   & 0.6190  \\
\bottomrule
\end{tabular}
\end{table}
To verify \oursm, we conduct bias evaluation experiments on three synthetic datasets, including SynFair, AttrBias, and StruBias. Specifically, for each dataset, we evaluate its bias using \oursm, and then calculate correlations between evaluated results and bias-controlling parameters, e.g., $|p_{00}-p_{01}|$ for SynFair, $\mu$ for AttrBias, and $p_{intra}$ for StruBias. As shown in Table~\ref{graddist_effect}, the evaluated results of \oursm present strong correlations with bias-controlling parameters. Specifically, across all three synthetic datasets, the Pearson correlation coefficients ($r$) consistently exceed $0.87$, peaking at $0.8983$ on the AttrBias dataset. This indicates a highly significant linear relationship, demonstrating that \oursm can proportionally capture the actual magnitude of the bias. Furthermore, the Spearman ($\rho$) and Kendall ($\tau$) coefficients, which measure rank correlation, also exhibit exceptional performance. Notably, on the AttrBias dataset, the Spearman and Kendall correlations reach $0.9643$ and $0.9048$, respectively. Even on the synthetically complex SynFair and StruBias datasets, the rank correlations remain robust ($\rho = 0.8214$, $\tau = 0.6190$). These high monotonic correlation scores suggest that \oursm is not only capable of quantifying the absolute severity of bias but is also highly reliable for ranking models or datasets according to their bias levels. Overall, these empirical results strongly validate the sensitivity and effectiveness of \oursm as a robust metric for bias evaluation without demographics.

\begin{table*}[]
\caption{Comparison of \ours using the GCN backbone with baseline methods on four datasets. In each row, the best result is marked in \textbf{bold}, while the runner-up result is marked with an \underline{underline}.
}
\centering
\label{tab:comparison_gcn}
\renewcommand\arraystretch{1.0}
\begin{tabular}{c|c|cccc|c}
\toprule
\textbf{Datasets} & \textbf{Metrics} & \textbf{Vanilla GCN}  & \textbf{FairGKD} & \textbf{Fairwos} & \textbf{FDKD} & \textbf{\ours} \\
\midrule
\multirow{4}{*}{\textbf{Bail}}                                   & F1 ($\uparrow$)                & 78.27 ± 0.89                   & \textbf{81.96 ± 1.05} & 77.59 ± 0.78          &    {\ul 79.50 ± 0.29}   & 78.89 ± 0.67    \\
                                   & ACC ($\uparrow$)               & 84.09 ± 0.73          &                \textbf{87.33 ± 0.78} & 83.35 ± 0.71          &   {\ul 85.17 ± 0.24}    & 83.97 ± 1.17    \\
                                   &  $\Delta_{DP}$ ($\downarrow$)           & {\ul 5.39 ± 0.28}  & 5.47 ± 0.27     & 5.98 ± 1.52           &  5.45 ± 0.24    & \textbf{4.35 ± 0.55}           \\
                                   
    &  $\Delta_{EO}$ ($\downarrow$)         & {\ul 3.21 ± 0.70}     & 3.72 ± 0.41           & 3.48 ± 1.72          &  3.57 ± 0.50     & \textbf{2.18 ± 1.08}  \\
    \midrule
\multirow{4}{*}{\textbf{Credit}}                                   & F1 ($\uparrow$)                & 82.13 ± 0.63          & 81.95 ± 0.11          & {\ul 82.87 ± 0.83}    &  81.90 ± 0.04   & \textbf{84.15 ± 2.23} \\
                                   & ACC ($\uparrow$)               & 73.81 ± 0.65          & 73.71 ± 0.13          & {\ul 74.54 ± 0.70} & 73.63 ± 0.05  & \textbf{75.03 ± 2.65}    \\
                                   &  $\Delta_{DP}$ ($\downarrow$)           & 11.93 ± 0.40          & 11.75 ± 0.11          & {\ul 10.47 ± 3.85}          & 11.56 ± 0.06    & \textbf{4.73 ± 4.63}  \\
                                   
  &  $\Delta_{EO}$ ($\downarrow$)         & 9.36 ± 0.24           & 9.49 ± 0.09           & {\ul 8.30 ± 3.34}           &   9.27 ± 0.08    & \textbf{3.64 ± 3.97}  \\
  \midrule
\multirow{4}{*}{\textbf{Pokec-z}}                                   & F1 ($\uparrow$)                & 70.23 ± 0.27    & 69.79 ± 0.48          & 69.88 ± 1.38          &  {\ul 70.53 ± 0.17}     & \textbf{70.70 ± 0.41}  \\
                                   & ACC ($\uparrow$)               & {\ul 69.74 ± 0.25} & 69.53 ± 0.19    & 69.09 ± 1.84          &  \textbf{69.94 ± 0.30}      & 68.85 ± 0.68          \\
                                   &  $\Delta_{DP}$ ($\downarrow$)           & 8.22 ± 0.48           &  {\ul 7.19 ± 0.81}     & 8.69 ± 2.49           &    8.07 ± 0.39   & \textbf{6.71 ± 0.45}  \\
 &  $\Delta_{EO}$ ($\downarrow$)         & 6.47 ± 0.56           & 6.02 ± 0.80     & 8.51 ± 1.94           &  {\ul 5.90 ± 0.61}     & \textbf{4.62 ± 0.49}  \\
 \midrule
\multirow{4}{*}{\textbf{Pokec-n}}                                   & F1 ($\uparrow$)                & 65.04 ± 0.33          & 64.88 ± 0.54          & \textbf{65.90 ± 1.85} & 65.42 ± 0.30  & {\ul 65.46 ± 1.64}    \\
                                   & ACC ($\uparrow$)               & 68.56 ± 0.36    & 68.31 ± 0.63          & \textbf{69.54 ± 1.13} & {\ul 69.14 ± 0.19}  & 67.20 ± 0.60          \\
                                   &  $\Delta_{DP}$ ($\downarrow$)           & 2.81 ± 0.79           & 2.73 ± 0.72           & 2.45 ± 0.80     &  {\ul 1.41 ± 0.60}   & \textbf{0.86 ± 0.44}  \\
 &  $\Delta_{EO}$ ($\downarrow$)         & 4.06 ± 0.92           & 4.18 ± 0.80           & {\ul 2.30 ± 1.84}     &  2.32 ± 0.79  & \textbf{2.22 ± 0.93} \\
\bottomrule
\end{tabular}
\end{table*}

\begin{table*}[]
\caption{Comparison of \ours using the GIN backbone with baseline methods on four datasets. In each row, the best result is marked in \textbf{bold}, while the runner-up result is marked with an \underline{underline}.
}
\centering
\label{tab:comparison_gin}
\renewcommand\arraystretch{1.0}
\begin{tabular}{c|c|cccc|c}
\toprule
\textbf{Datasets} & \textbf{Metrics} & \textbf{Vanilla GIN}  & \textbf{FairGKD} & \textbf{Fairwos} & \textbf{FDKD} & \textbf{\ours} \\
\midrule
\multirow{4}{*}{\textbf{Bail}} & F1 ($\uparrow$)               & 72.04 ± 2.18          & {\ul 75.29 ± 3.47}    & \textbf{75.50 ± 0.57} &   68.06 ± 2.86      & 68.99 ± 5.92          \\
                                   & ACC ($\uparrow$)              & 77.20 ± 2.51          & \textbf{81.10 ± 3.78}  & {\ul 80.80 ± 1.62}     &      72.66 ± 2.87         & 77.40 ± 4.52          \\
                                   & $\Delta_{DP}$ ($\downarrow$)           & 7.79 ± 0.80           & \textbf{4.73 ± 1.42}  & 6.29 ± 0.75           &    5.77 ± 1.05    & {\ul 4.87 ± 2.38}     \\
                                   &  $\Delta_{EO}$ ($\downarrow$)         & 6.55 ± 0.87           & {\ul 3.39 ± 1.93}     & 3.62 ± 0.47           &  4.30 ± 1.81   & \textbf{3.18 ± 1.76}  \\
                                   \midrule
\multirow{4}{*}{\textbf{Credit}} & F1 ($\uparrow$)               & 82.82 ± 0.46          & {\ul 83.01 ± 0.61}    & 80.94 ± 2.07          &      82.24 ± 0.41           & \textbf{83.25 ± 1.14} \\
                                   & ACC ($\uparrow$)              & 74.61 ± 0.51          & \textbf{74.85 ± 0.71} & 72.45 ± 2.40          &       73.98 ± 0.47          & {\ul 74.84 ± 1.30}    \\
                                   & $\Delta_{DP}$ ($\downarrow$)           & 11.31 ± 1.06          & {\ul 10.62 ± 1.92}    & 10.98 ± 2.43          &    12.05 ± 1.23             & \textbf{4.85 ± 3.30}  \\
                                   &  $\Delta_{EO}$ ($\downarrow$)         & 8.77 ± 1.20           & 8.32 ± 1.61           & {\ul 8.30 ± 2.88}     &    9.46 ± 1.03             & \textbf{2.47 ± 2.49}  \\
                                   \midrule
\multirow{4}{*}{\textbf{Pokec-z}} & F1 ($\uparrow$)               & 67.64 ± 0.91    & \textbf{68.84 ± 1.17} & 67.14 ± 2.06          &     {\ul 68.82 ± 0.86}            & 67.33 ± 1.54          \\
                                   & ACC ($\uparrow$)              & 68.30 ± 0.98    & 67.97 ± 1.06          & {\ul 68.87 ± 0.75} &  \textbf{68.99 ± 0.68}             & 66.66 ± 2.28          \\
                                   & $\Delta_{DP}$ ($\downarrow$)           & 3.57 ± 0.95           & {\ul 1.79 ± 0.68}     & 6.46 ± 4.36           &      3.68 ± 1.03           & \textbf{1.64 ± 1.20}  \\
                                   &  $\Delta_{EO}$ ($\downarrow$)         & 4.75 ± 0.85           & {\ul 1.81 ± 1.15}     & 6.09 ± 3.68           &  3.90 ± 0.76                 & \textbf{1.72 ± 1.48}  \\
                                   \midrule
\multirow{4}{*}{\textbf{Pokec-n}} & F1 ($\uparrow$)               & 62.67 ± 1.01          & {\ul 63.27 ± 0.94}    & \textbf{64.50 ± 1.59} &      61.75 ± 0.98           & {\ul 63.27 ± 0.81}    \\
                                   & ACC ($\uparrow$)              & 67.87 ± 0.91          & 67.40 ± 0.65          & \textbf{69.79 ± 1.00} &  {\ul 68.76 ± 0.50}               & 68.10 ± 0.83    \\
                                   & $\Delta_{DP}$ ($\downarrow$)           & {\ul 0.78 ± 0.78}     & 2.33 ± 1.39           & 2.98 ± 1.37           &      0.80 ± 0.60           & \textbf{0.63 ± 0.32}  \\
                                   &  $\Delta_{EO}$ ($\downarrow$)         & 3.37 ± 1.24     & 3.77 ± 1.74           & 5.85 ± 1.46           &    {\ul 3.18 ± 1.17}              & \textbf{2.47 ± 0.86} \\
                                   \bottomrule
\end{tabular}
\end{table*}

\subsubsection{Comparison Study}
\label{subsubsec:comparison}
To evaluate the effectiveness of \ours, we compare it against three state-of-the-art fairness methods using two standard GNN backbones. Tables~\ref{tab:comparison_gcn} and~\ref{tab:comparison_gin} summarize the performance comparisons on the GCN and GIN backbones, respectively. In most cases, \ours achieves superior utility and fairness trade-offs compared to the baselines. This observation demonstrates the efficacy of \ours in improving graph fairness without relying on demographic information. Furthermore, the consistent performance across different GNN architectures highlights the model-agnostic nature of \ours.

It is worth noting that \ours experiences a significant drop in F1 score on the Bail dataset when using the GIN backbone. This can be attributed to the architectural properties of GIN and the underlying structure of the dataset. Specifically, while GIN is skilled at capturing local topological features compared to GCN, the edges in the Bail dataset are constructed based on node feature similarity via kNN and lack the rich structural complexity found in social networks, e.g., Pokec-z/n. Consequently, a potential explanation is that the bias amplification stage of \ours struggles to effectively amplify biases on the Bail dataset under the GIN backbone, as high-confidence and low-confidence samples share similar local topologies. We provide a more detailed analysis of the bias amplification stage in Section~\ref{subsec:ablation}.

Furthermore, we observe that while FDKD successfully improves fairness and preserves utility on the GCN backbone, it struggles to achieve comparable effectiveness with GIN on the Bail dataset. As a fairness baseline not explicitly tailored for graph data, FDKD employs label smoothing to improve fairness, which acts as a reweighting strategy that assigns higher importance to correctly classified samples. Since this mechanism is conceptually similar to the bias amplification stage of \ours, we attribute this underperformance to the same underlying factor that limits \ours when using the GIN backbone on the Bail dataset.

\subsection{Ablation Study}
\label{subsec:ablation}
\begin{figure}[!t]
  \centering
  \includegraphics[width=\linewidth]{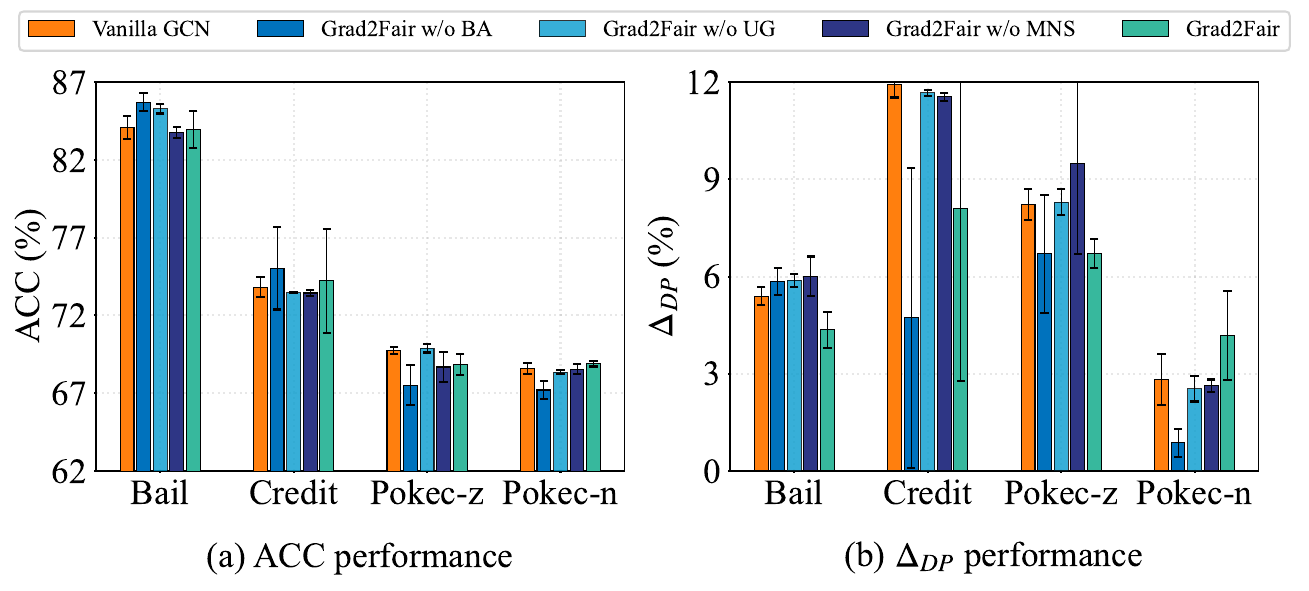}
  \caption{Ablation results of \ours using the GCN backbone.}
  \label{fig:ablation_gcn}
\end{figure}

\begin{figure}[!t]
  \centering
  \includegraphics[width=\linewidth]{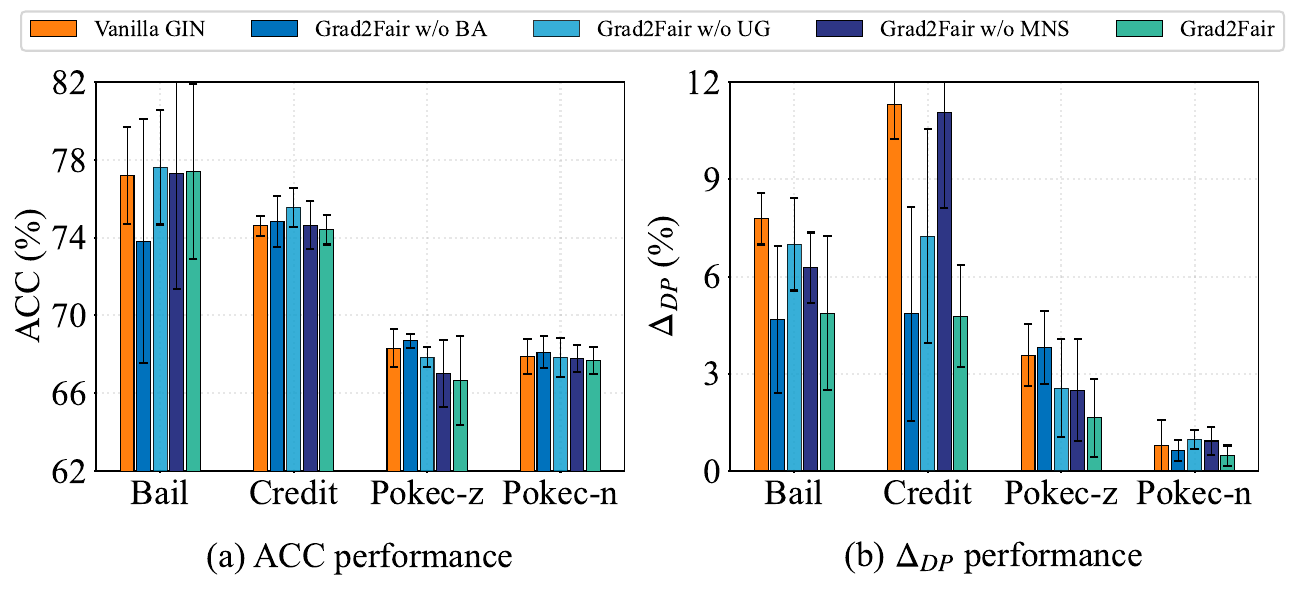}
  \caption{Ablation results of \ours using the GIN backbone.}
  \label{fig:ablation_gin}
\end{figure}
To evaluate the contributions of the core components in \ours, we conduct ablation studies on two GNN backbones. Specifically, we remove the bias amplification and upweighting via gradient stages, denoting these variants as ``\ours w/o BA" and ``\ours w/o UG", respectively. Additionally, we replace upweighting for the misclassified node in Eq.~\eqref{eq:upweight} with upweighting across all training nodes, denoted as ``\ours w/o MNS". 

Figures~\ref{fig:ablation_gcn} and \ref{fig:ablation_gin} present the ablation results on the GCN and GIN backbones, respectively. In most cases, removing the bias amplification (BA) stage improves fairness, validating its intended function of magnifying inherent biases. Notably, for datasets with severe inherent bias, such as Credit, removing the BA stage enhances both fairness and utility. Consequently, the performance of \ours on the Credit and Pokec-n datasets (Tables~\ref{tab:comparison_gcn} and~\ref{tab:comparison_gin}) is reported without the BA stage. Furthermore, on the Bail dataset, the removal of the BA stage yields contrasting results between the two backbones. We attribute this discrepancy to the simple topology of the Bail dataset and GIN's superior ability to capture local structural patterns. As a result, during bias amplification, \ours equipped with GIN struggles to distinguish the local topological differences between high- and low-confidence samples, thereby rendering the BA stage ineffective.

Regarding the upweighting via gradient (UG) stage, removing it degrades the model's fairness to a level comparable with the vanilla baseline. This demonstrates that the UG stage effectively mitigates model bias and enhances fairness. By jointly analyzing the results of ``\ours w/o BA" and ``\ours w/o UG", we find that the BA stage is less effective on artificially constructed graphs (e.g., Bail and Credit), largely due to their lack of complex local topological structures. Finally, the results of ``\ours w/o MNS" demonstrate that applying gradient upweighting uniformly to all nodes fails to yield consistent fairness improvements and can even exacerbate bias. This finding validates the effectiveness and rationale of restricting gradient upweighting exclusively to misclassified samples, as formulated in Eq.~\eqref{eq:upweight}.

\subsection{Hyperparameter Analysis}
To analyze the impact of key hyperparameters, we investigate the performance sensitivity of \ours on the selection ratio $\tau$ and debiasing strength $\lambda$.

\subsubsection{Sensitivity w.r.t. $\tau$}
\begin{figure}[!t]
  \centering
  \includegraphics[width=\linewidth]{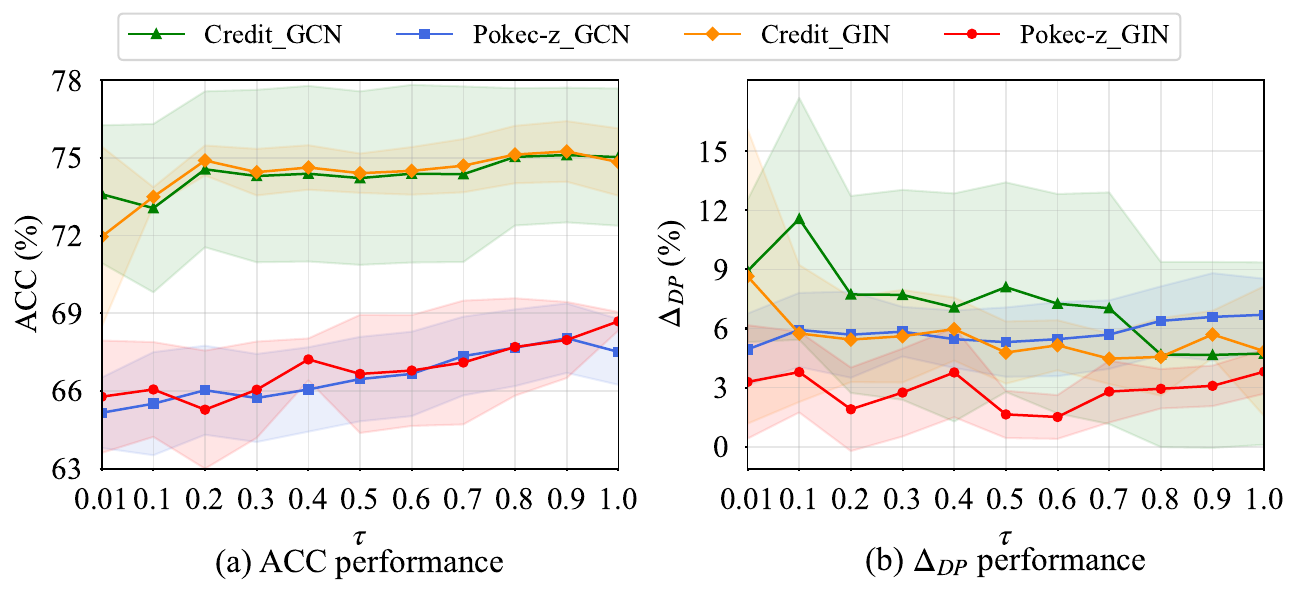}
  \caption{Hyperparameter sensitivity of $\tau$ on the Credit and Pokec-z datasets.}
  \label{fig:para_sens_tau}
\end{figure}

We vary $\tau$ across the set $\{0.01, 0.1, 0.2, 0.3, 0.4, 0.5, 0.6, 0.7, 0.8, 0.9, 1.0\}$ while keeping all other parameters identical to those in Section~\ref{subsubsec:comparison}. As shown in Figure~\ref{fig:para_sens_tau}, we observe that the utility performance of \ours improves as $\tau$ increases, while the fairness performance remains stable. Since the BA stage only selects a subset to optimize the model, as $\tau$ increases, the training of the BA stage increasingly approximates standard model training, thereby enhancing utility. Meanwhile, the stable fairness performance across a broad range of variations in $\tau$ demonstrates the effectiveness of the UG stage on debiasing.

\subsubsection{Sensitivity w.r.t. $\lambda$}
\begin{figure}[!t]
  \centering
  \includegraphics[width=\linewidth]{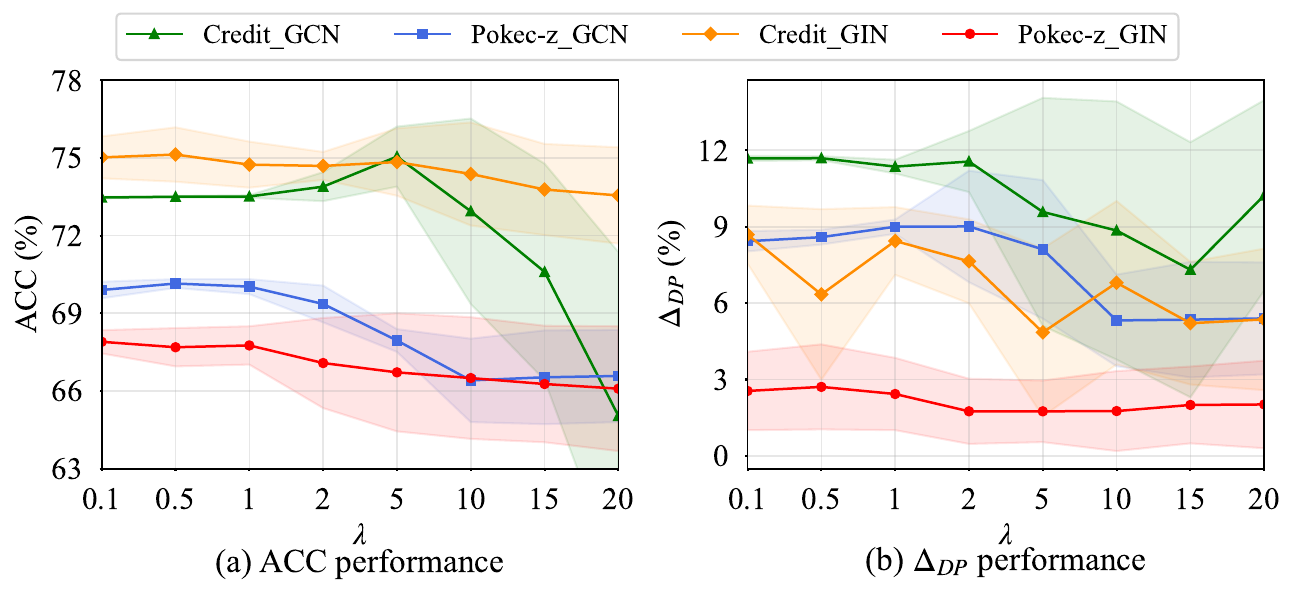}
  \caption{Hyperparameter sensitivity of $\lambda$ on the Credit and Pokec-z datasets.}
  \label{fig:para_sens_lambda}
\end{figure}

We vary $\lambda$ across the set $\{0.1, 0.5, 1, 2, 5, 10, 15, 20\}$ while keeping all other parameters identical to those in Section~\ref{subsubsec:comparison}. As shown in Figure~\ref{fig:para_sens_lambda}, fairness improves at the expense of utility as $\lambda$ increases. Despite this inherent trade-off, the overall performance of \ours remains robust to variations in $\lambda$. Functionally, $\lambda$ controls the extent to which the model learns from misclassified samples. While increasing $\lambda$ forces the model to prioritize these challenging samples and enhances fairness, it inevitably compromises utility performance. Therefore, selecting an optimal $\lambda$ requires carefully balancing the fairness-utility trade-off.

\subsection{Further Probe}
To obtain an in-depth understanding of \ours, we investigate the training efficiency, gradient visualization, and embedding visualization.

\subsubsection{Training Efficiency}
\begin{figure}[!t]
  \centering
  \includegraphics[width=\linewidth]{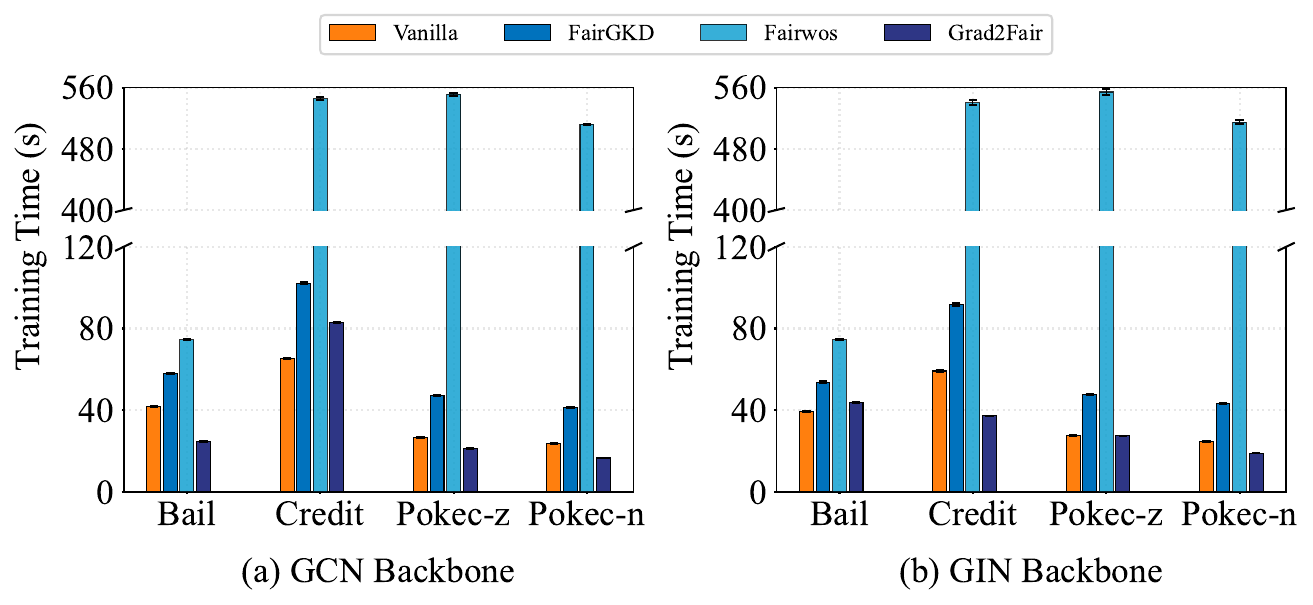}
  \caption{Comparison of training times based on the GCN and GIN backbones. The best result is marked in \textbf{bold}.}
  \label{fig:training_time}
\end{figure}

We compare the training time of \ours with that of several baselines. We maintain hyperparameter settings consistent with those in Section~\ref{subsubsec:comparison}. As shown in Figure~\ref{fig:training_time}, we report the total training time averaged over five random seeds. The training time of \ours is markedly lower than that of the two fairness baselines (i.e., FairGKD and Fairwos), which demonstrates the high efficiency of \ours. Compared to these two baselines, \ours mitigates model bias through upweighting, which incurs no additional computational overhead and strictly matches the per-epoch cost of training vanilla models. Additionally, \ours exhibits higher efficiency than the vanilla models in most cases, which can be attributed to fewer required training epochs of \ours.

To ensure a fair comparison independent of the total number of epochs required for convergence, we further analyze the per-epoch training time. As shown in Tables~\ref{tab:training_time_gcn} and~\ref{tab:training_time_gin}, \ours achieves a lower per-epoch training time than both FairGKD and Fairwos. This indicates that the reduced total training time of \ours (shown in Figure~\ref{fig:training_time}) stems not only from requiring fewer epochs to converge but also from a lower computational cost per epoch. Furthermore, \ours maintains a per-epoch training time comparable to the vanilla baseline, demonstrating that our approach introduces negligible additional computational overhead. Overall, these empirical results confirm that \ours is a highly efficient fairness method.
\begin{table}[!t]
\centering
\caption{Comparison of training times (in seconds) using the GCN backbone. The best result is marked in \textbf{bold}.}
\label{tab:training_time_gcn}
\renewcommand\arraystretch{1.0}
\resizebox{\linewidth}{!}{
    \begin{tabular}{ccccc}
        \toprule
        \textbf{Datasets} & \textbf{Vanilla} & \textbf{FairGKD} & \textbf{Fairwos} & \textbf{\ours} \\
        \midrule
        \textbf{Bail}     & 0.0411  & 0.0437  & 0.0611  & \textbf{0.0407}    \\
        \textbf{Credit}   & 0.0642  & 0.0753  & 0.4402  & \textbf{0.0631}    \\
        \textbf{Pokec-z}  & 0.0261  & 0.0294  & 0.4530  & \textbf{0.0261}    \\
        \textbf{Pokec-n}  & \textbf{0.0229}  & 0.0258  & 0.4263  & 0.0231    \\
        \bottomrule
    \end{tabular}}
\end{table}

\begin{table}[!t]
\centering
\caption{Comparison of training times (in seconds) using the GIN backbone. The best result is marked in \textbf{bold}.}
\label{tab:training_time_gin}
\renewcommand\arraystretch{1.0}
\resizebox{\linewidth}{!}{
    \begin{tabular}{ccccc}
        \toprule
        \textbf{Datasets} & \textbf{Vanilla} & \textbf{FairGKD} & \textbf{Fairwos} & \textbf{\ours} \\
        \midrule
        \textbf{Bail}     & \textbf{0.0390}  & 0.0412  & 0.0615  & 0.0397    \\
        \textbf{Credit}   & \textbf{0.0594}  & 0.0720  & 0.4435  & 0.0626    \\
        \textbf{Pokec-z}  & \textbf{0.0270}  & 0.0303  & 0.4582  & 0.0287    \\
        \textbf{Pokec-n}  & \textbf{0.0242}  & 0.0267  & 0.4253  & 0.0272    \\
        \bottomrule
    \end{tabular}}
\end{table}

\subsubsection{Gradient Visualization}
\begin{figure}[!t]
  \centering
  \includegraphics[width=\linewidth]{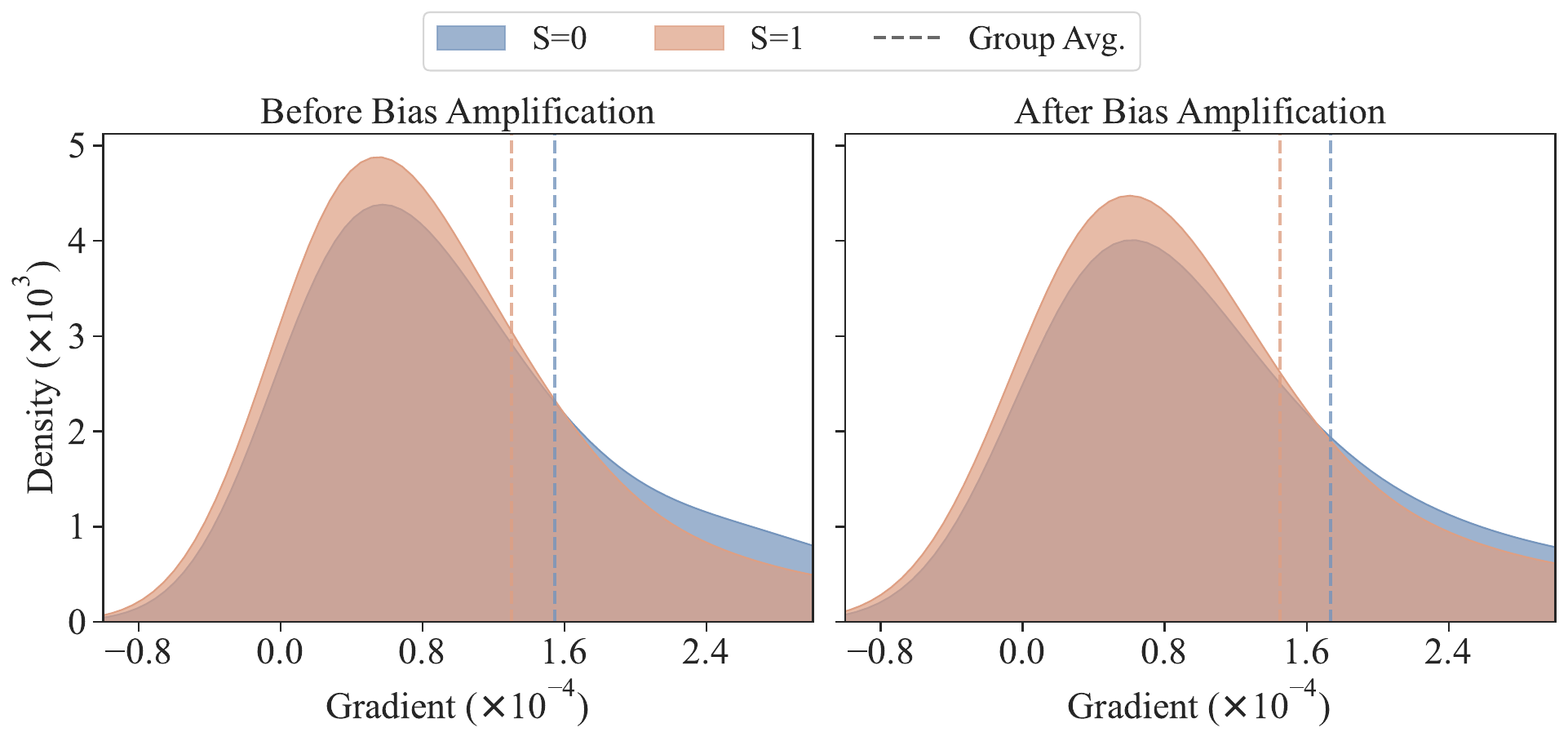}
  \caption{Visualization of gradients before and after bias amplification using a GCN backbone on the Pokec-z dataset.}
  \label{fig:grad_visual_gcn}
\end{figure}

\begin{figure}[!t]
  \centering
  \includegraphics[width=\linewidth]{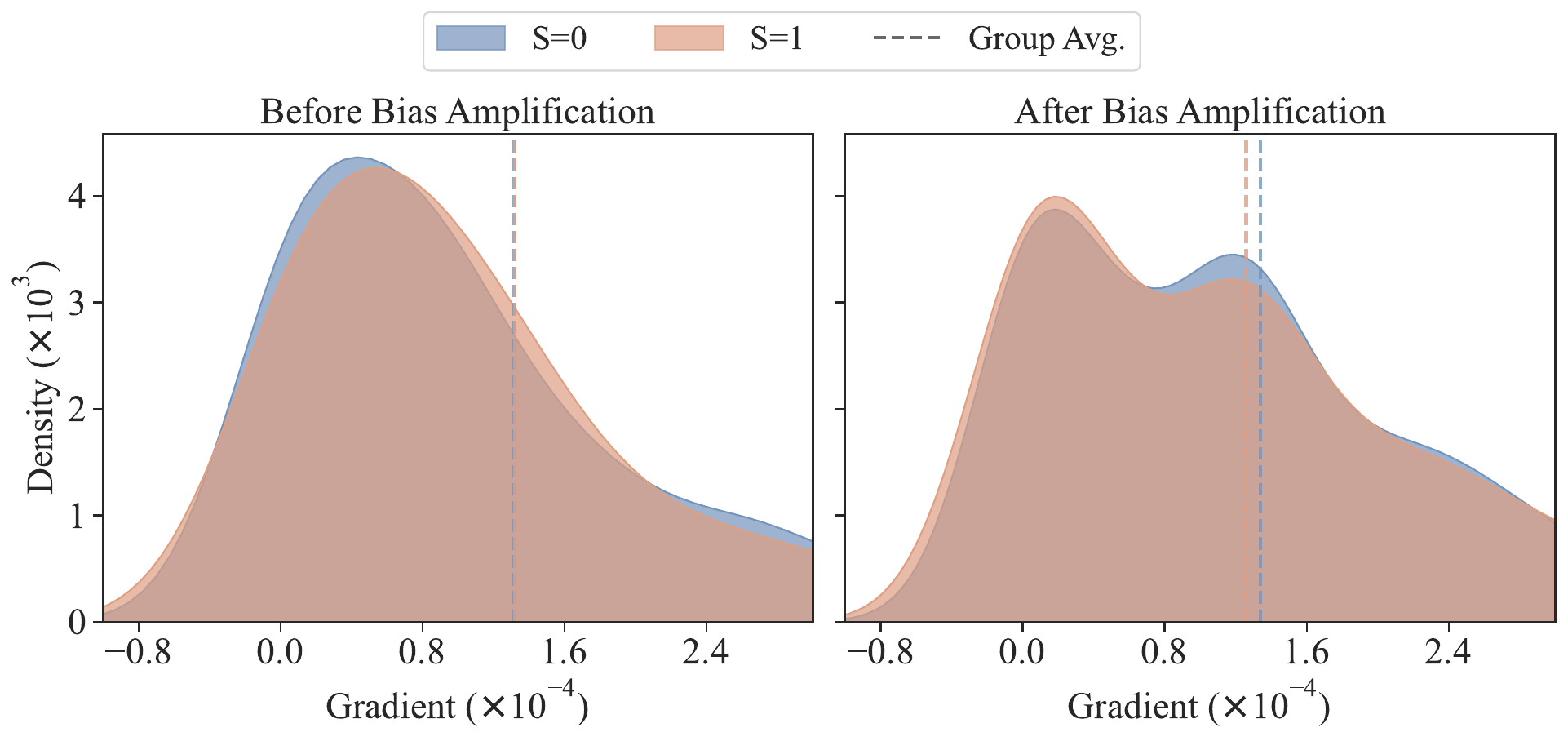}
  \caption{Visualization of gradients before and after bias amplification using a GIN backbone on the Pokec-z dataset.}
  \label{fig:grad_visual_gin}
\end{figure}

To further investigate the impact of the BA stage, we visualize the gradient distribution before and after the BA stage over two GNN backbones. Specifically, we present gradients of training nodes misclassified by the model after bias amplification. As shown in Figures~\ref{fig:grad_visual_gcn} and~\ref{fig:grad_visual_gin}, while bias amplification induces marginal changes in the overall gradient distributions across different groups, an analysis of the average gradients reveals a widening gap between these groups after bias amplification. This can be observed by the distance between the two dashed lines in Figures~\ref{fig:grad_visual_gcn} and~\ref{fig:grad_visual_gin}. The discrepancy in average gradients on the GCN backbone increases from $2.42 \times 10^{-5}$ to $2.85 \times 10^{-5}$. More notably, on the GIN backbone, this difference increases from $6 \times 10^{-7}$ to $7.9 \times 10^{-6}$, representing a full order-of-magnitude increase. Overall, this increasing distance transforms meaningless gradients into demographic information for debiasing. 

\subsubsection{Embedding Visualization}
\begin{figure}[!t]
    \centering
    \begin{minipage}[b]{0.48\linewidth}
        \centering
        \includegraphics[width=\linewidth]{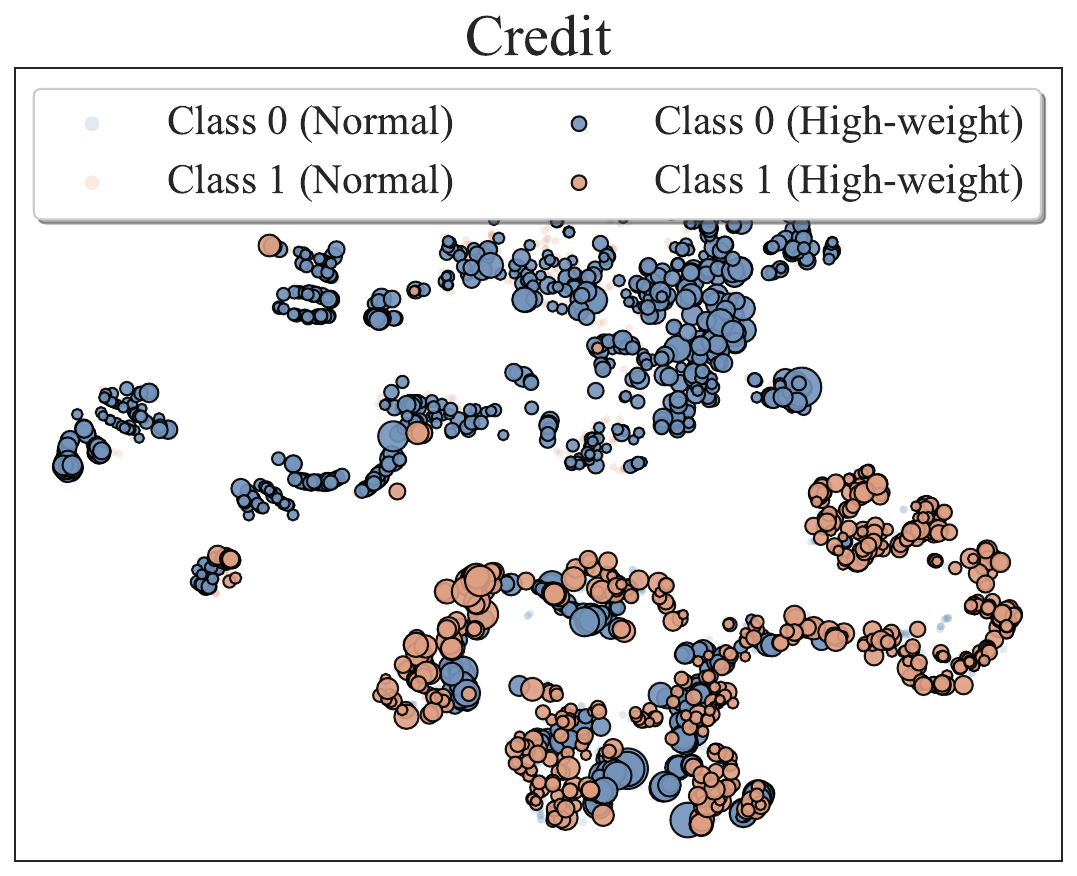}
    \end{minipage}
    \hfill 
    \begin{minipage}[b]{0.48\linewidth}
        \centering
        \includegraphics[width=\linewidth]{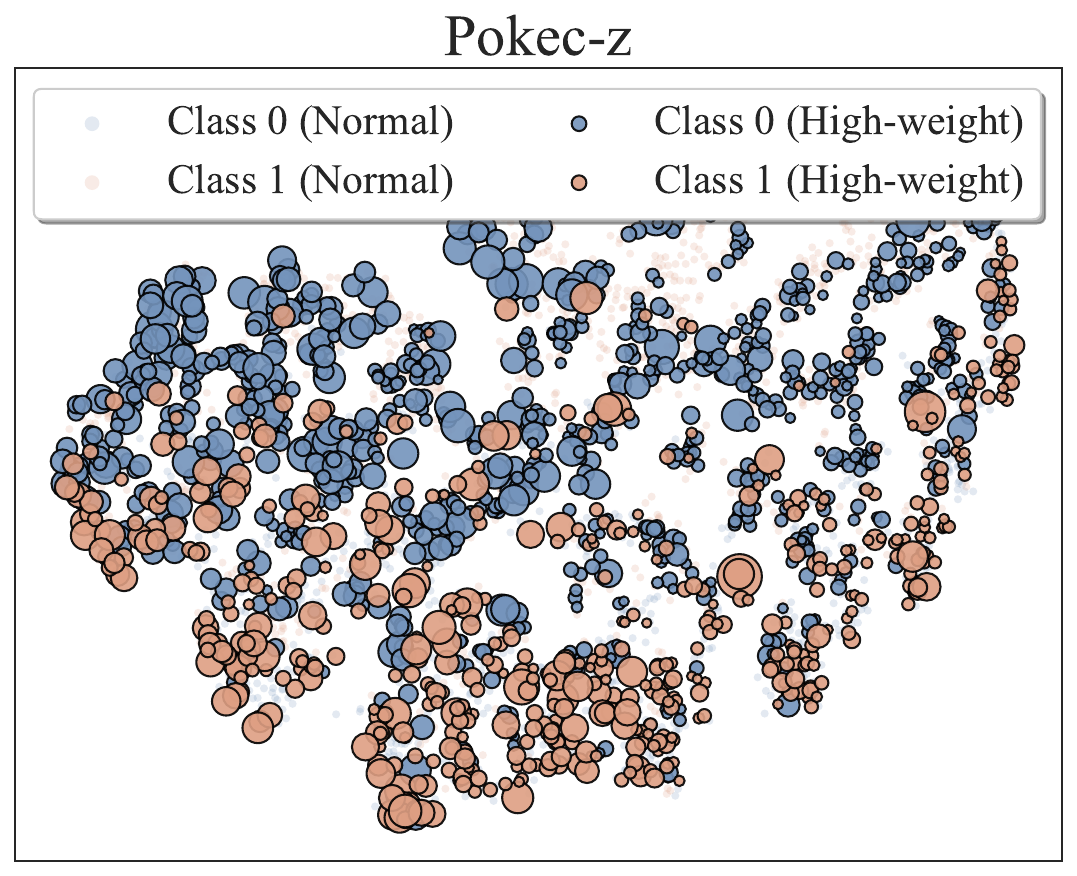}
    \end{minipage}
    
    \caption{T-SNE visualization of embedding of \ours using the GCN backbone on the Credit and Pokec-z datasets.}
    \label{fig:tsne}
\end{figure}
As shown in Figure~\ref{fig:tsne}, we visualize the node embeddings generated by \ours using t-SNE, with nodes color-coded according to their binary labels. To illustrate the spatial distribution of weights, normal nodes (those not in $\mathcal{M}$) are rendered as semi-transparent points, while high-weight nodes (those in $\mathcal{M}$) are emphasized as opaque markers with black contours, their sizes scaled proportionally to their weight magnitude. Although all nodes are naturally partitioned into two distinct clusters, the spatial distributions of normal and high-weight nodes are strikingly inverted relative to their labels. Specifically, while normal nodes seamlessly align with their true classes, the high-weight nodes of one class structurally overlap with the normal nodes of the opposing class. For instance, the normal nodes of class 0 and the high-weight nodes of class 1 are densely superimposed. Crucially, these high-weight nodes are conspicuously absent from the homophilous cores of their own ground-truth clusters. Instead, they are either densely concentrated along the inter-class decision boundary or scattered as isolated, topologically heterophilous instances deep within the opposing cluster. This observation demonstrates that \ours is capable of accurately locating the decision boundary, thereby precisely identifying the most vulnerable victim nodes without relying on demographic labels $S$.

\section{Conclusion}
\label{sec:conclusion}
In this paper, we investigate the problem of graph fairness in the absence of demographic information. We first observe that the gradients of misclassified nodes encode latent demographic signals. Motivated by this observation, we introduce \oursm, a novel bias evaluation metric designed for demographics-free scenarios. \oursm quantifies bias by measuring the distance between local modes within the gradient distribution. Consequently, we propose \ours, a simple yet effective framework for demographics-free graph fairness. \ours adopts an amplify-then-debias paradigm, comprising two key stages: a bias amplification and an upweighting via gradient stage. The core intuition behind \ours is to exploit the underlying demographic information embedded within the gradients to mitigate bias. To validate the soundness of our approach, we provide a rigorous theoretical analysis of both stages. Extensive experiments on several datasets further demonstrate the empirical efficacy of \oursm and \ours.

Despite the strong empirical and theoretical results, our approach has certain limitations. First, \oursm is currently restricted to evaluating bias with respect to binary sensitive attributes, whereas many real-world applications involve multi-valued sensitive attributes, such as race or geographic region. Second, on datasets characterized by simplistic topological structures, the BA stage of \ours may exhibit reduced efficacy. Under such conditions, extracting demographic signals becomes challenging, which ultimately compromises the overall fairness improvements.

\begin{IEEEbiography}[{\includegraphics[width=1in,height=1.25in,clip,keepaspectratio]{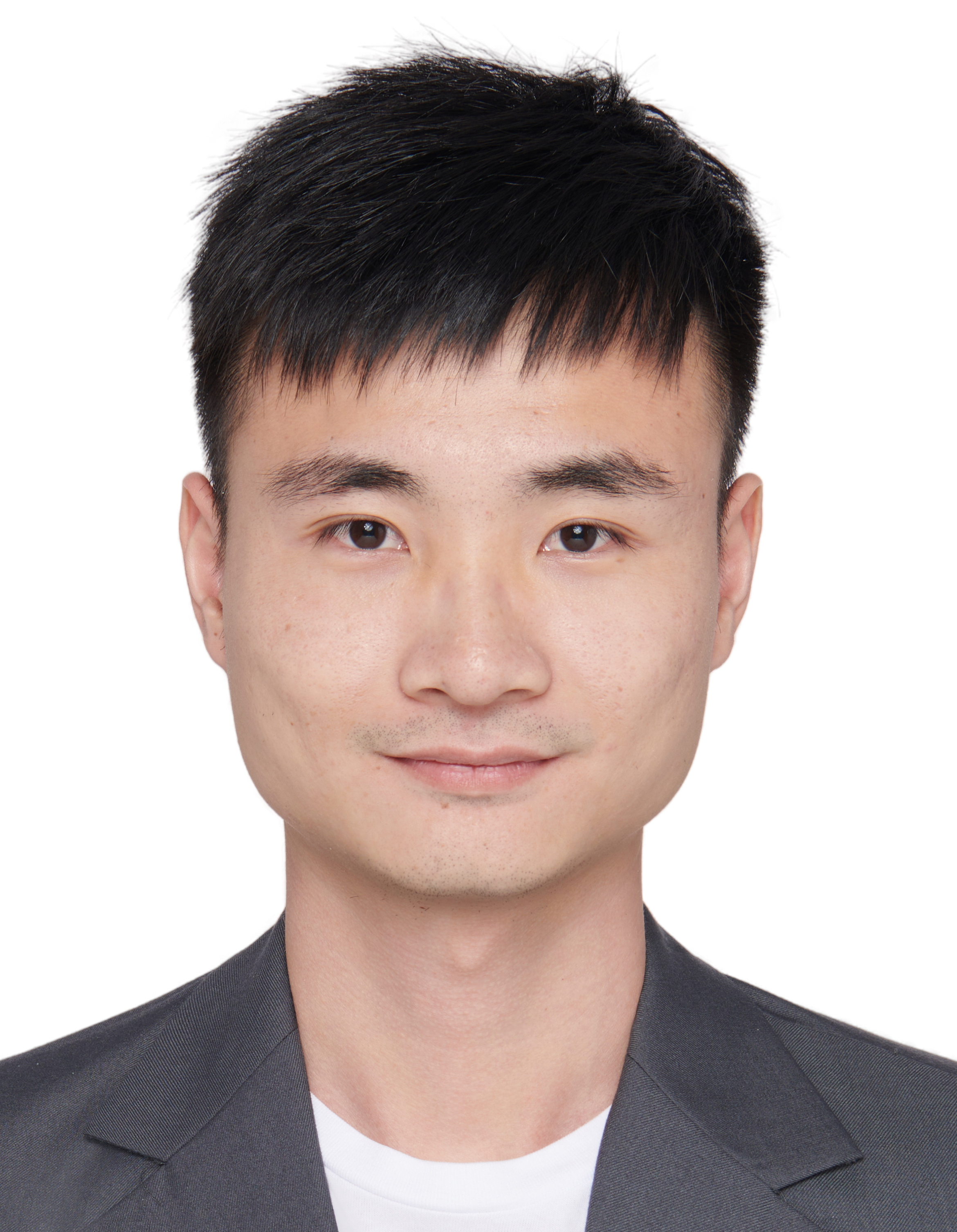}}]{Yuchang Zhu}
is a Ph.D. student at the School of Computer Science and Engineering, Sun Yat-sen University, Guangzhou, China. He received the master's degree from the College of Engineering, South China Agricultural University. His main research interests include graph machine learning, graph fairness, and data mining techniques. Over the past three years, he has published papers in top journals/conferences, including TKDE, TPAMI, ICML, KDD, WWW, and AAAI.
\end{IEEEbiography}


\begin{IEEEbiography}[{\includegraphics[width=1in,height=1.25in,clip,keepaspectratio]{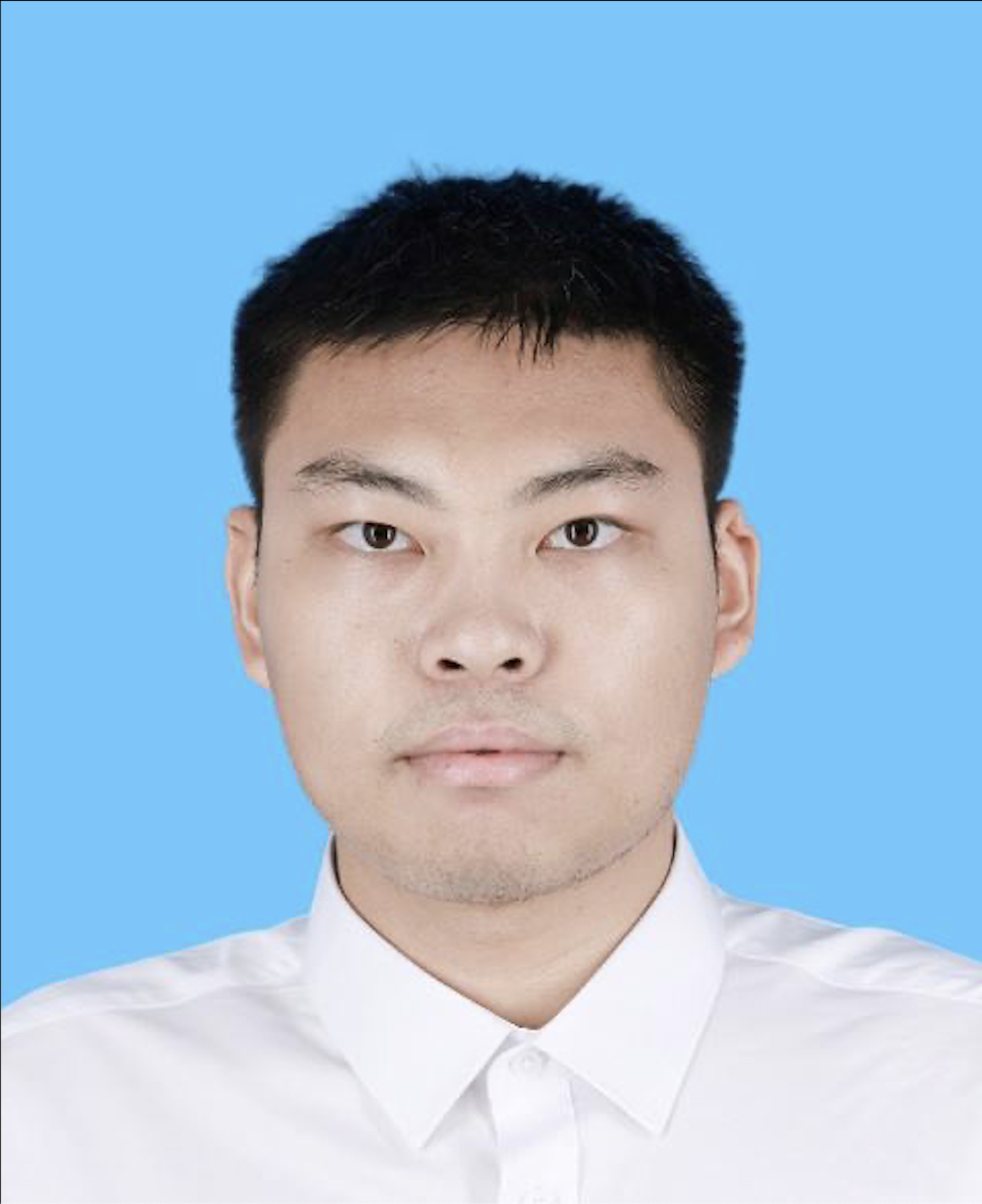}}]{Zezhong Xie} is an undergraduate student at Sun Yat-sen University. His research focuses on generative recommendations and their fairness. 
\end{IEEEbiography}


\begin{IEEEbiography}
[{\includegraphics[width=1in,height=1.25in,clip,keepaspectratio]{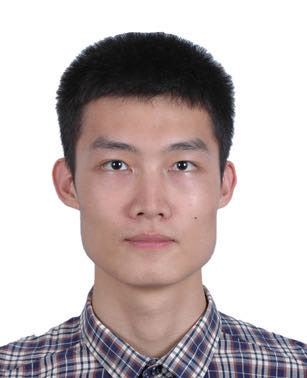}}]{Huizhe Zhang} is a Ph.D. student at the School of Computer Science and Engineering, Sun Yat-sen University, Guangzhou, China. He received the master's degree from the school of computer science and technology, Guangdong University of Technology. His main research interests include spiking neural networks and graph data mining techniques.
\end{IEEEbiography}

\begin{IEEEbiography}
[{\includegraphics[width=1in,height=1.25in,clip,keepaspectratio]{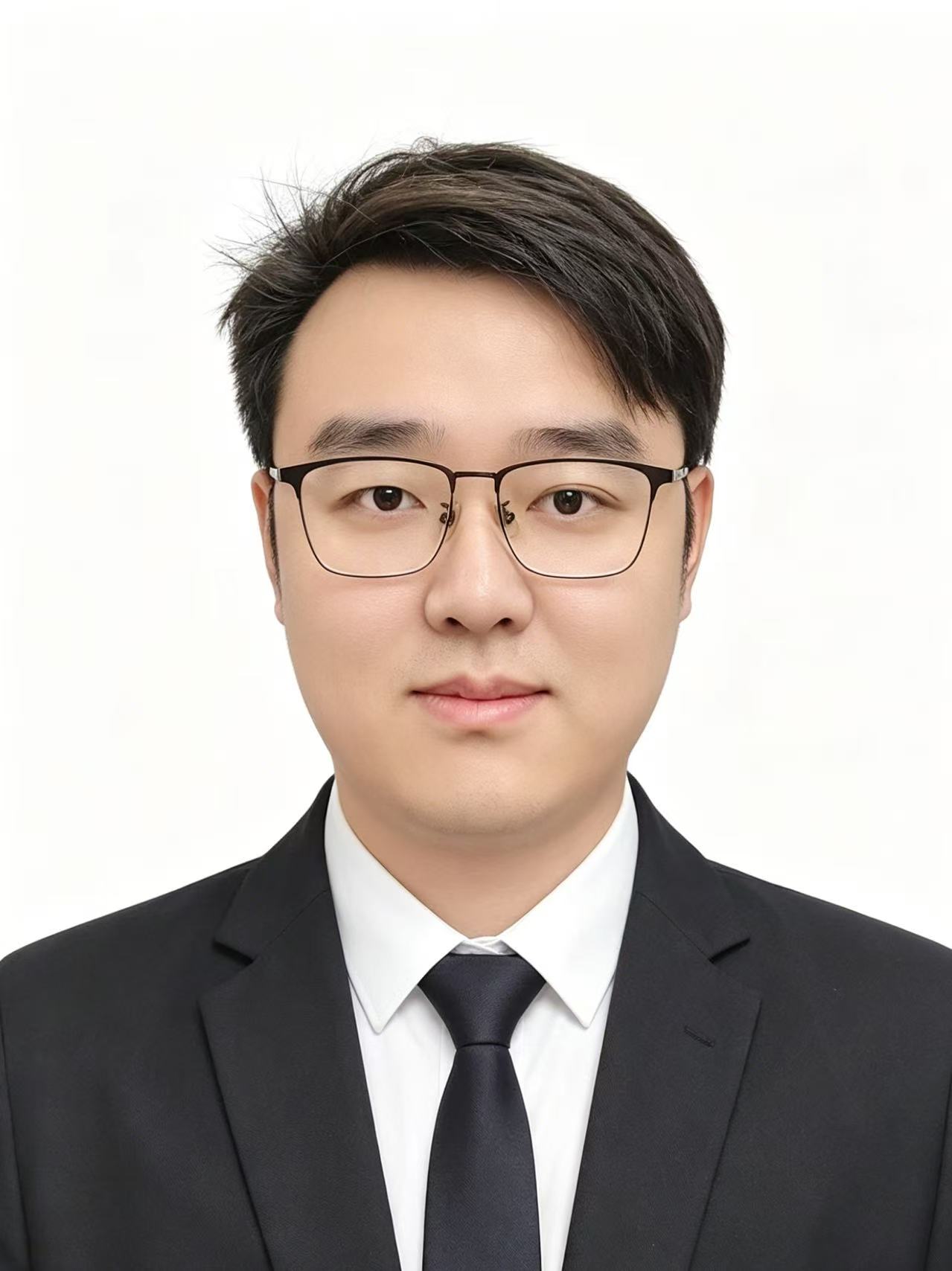}}]{Huazhen Zhong} is a Ph.D. student at the School of Computer Science and Engineering, Sun Yat-sen University, Guangzhou, China. He received the master's degree from the Institute of Information Engineering, Chinese Academy of Sciences, Beijing, China. His main research interests include multimodal learning, and recommendation systems.
\end{IEEEbiography}

\begin{IEEEbiography}[{\includegraphics[width=1in,height=1.25in,clip,keepaspectratio]{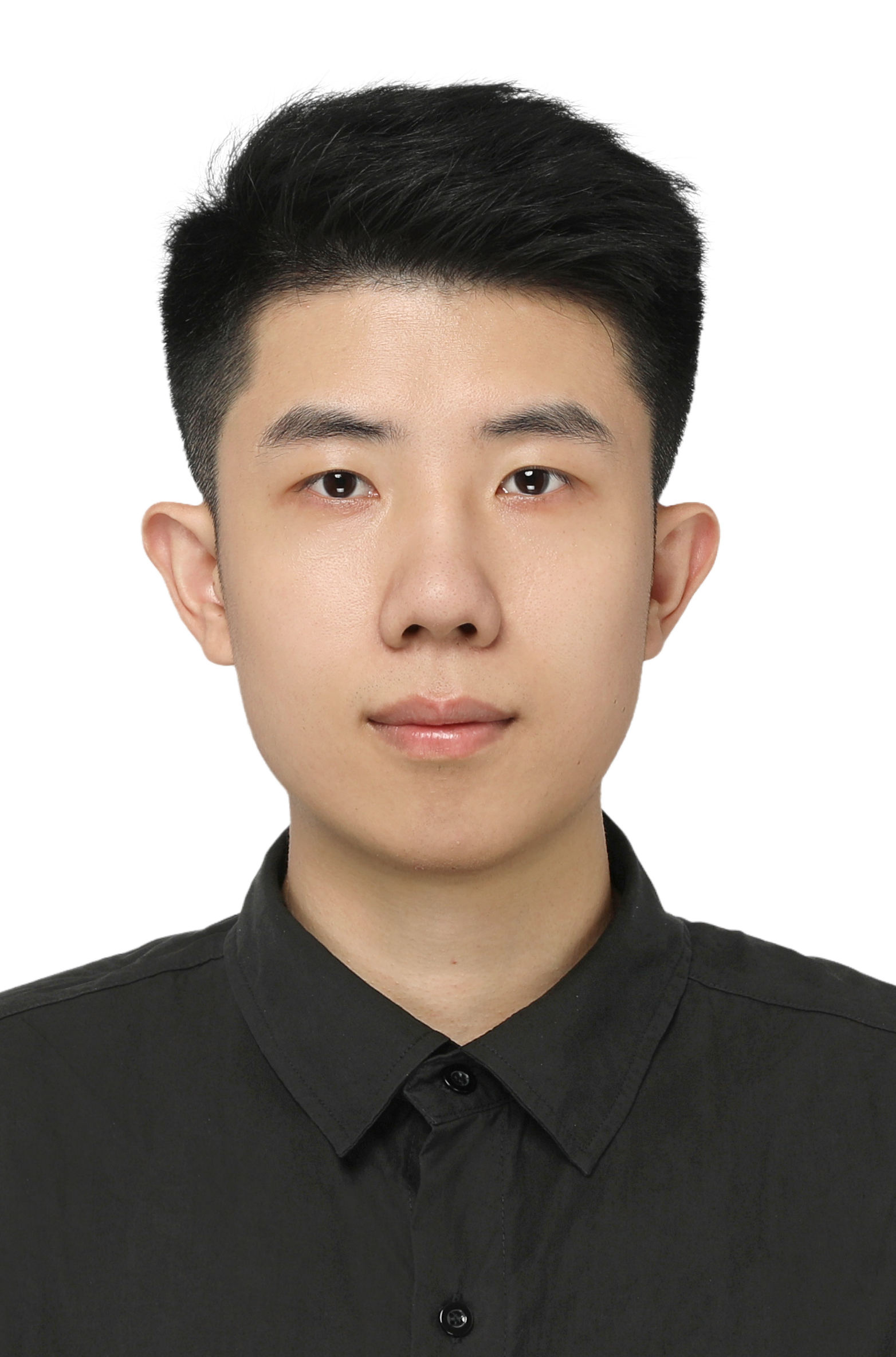}}]{Jintang Li} is currently an assistant professor at the Institute of Artificial Intelligence, Xiamen University, China. He received his M.S. and Ph.D. degrees from Sun Yat-sen University in 2021 and 2025, respectively. His research interests focus on graph representation learning and data mining. Over the past five years, he has published more than 30 papers in leading journals and conferences, including TPAMI, TKDE, KDD, AAAI, WWW, CIKM, WSDM, ICLR, NeurIPS, IJCAI, and ICML. He received the ACM China (Zhuhai Chapter) Doctoral Dissertation Award in 2025. He also serves as a program committee member for several top conferences, such as NeurIPS, ICLR, WWW, KDD, IJCAI, AAAI, and CIKM, and as a regular reviewer for journals including TKDE, TKDD, and TNNLS.
\end{IEEEbiography}


\begin{IEEEbiography}[{\includegraphics[width=1in,height=1.25in,clip,keepaspectratio]{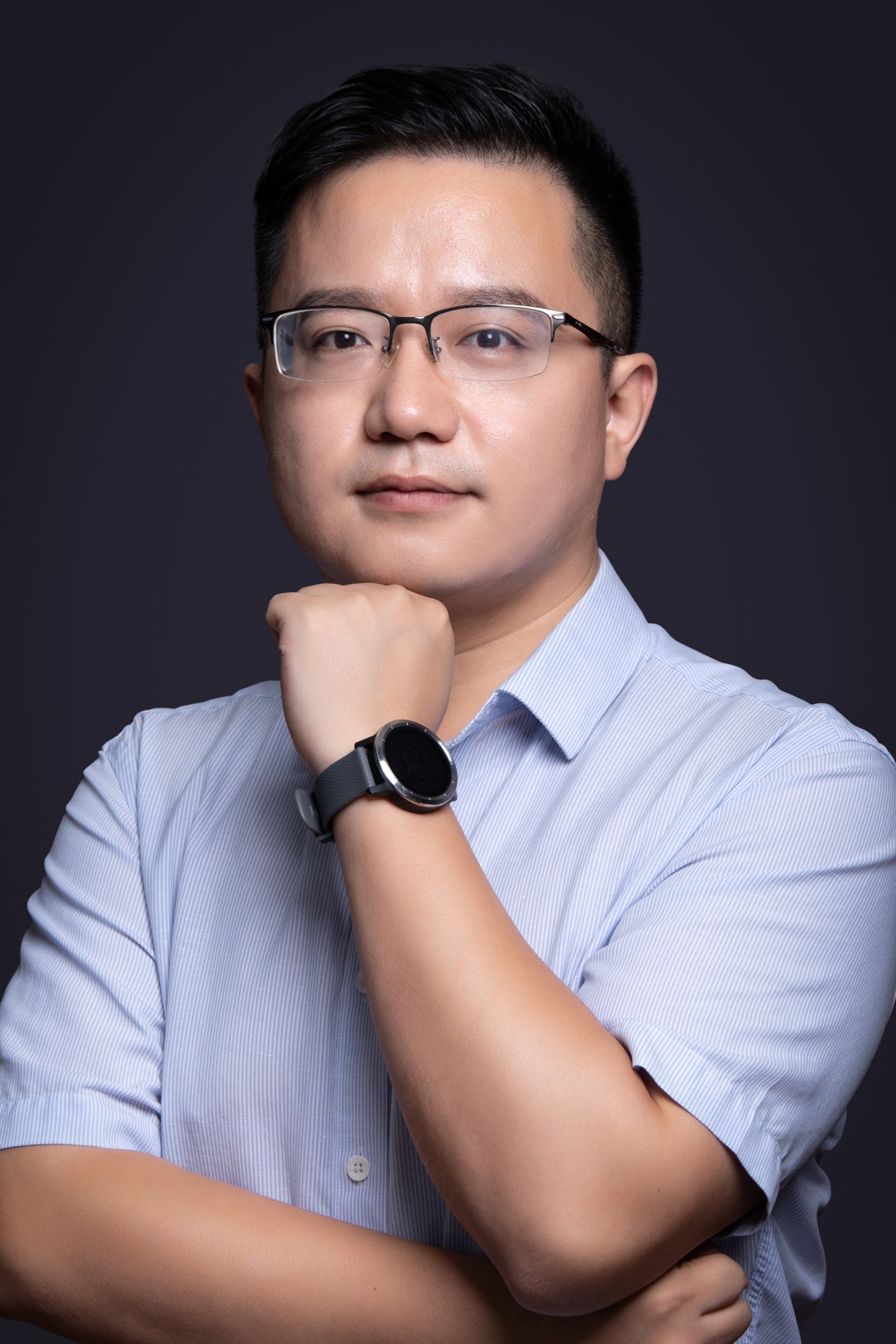}}]{Liang Chen} is currently an associate professor with the School of Computer Science and Engineering, Sun Yat-Sen University (SYSU), China. He received the bachelor’s and Ph.D. degrees from Zhejiang University (ZJU) in 2009 and 2015, respectively. His research areas include trustworthy machine learning, Large Language Models, and data mining. In the recent five years, he has published over 80 papers in several top conferences/journals, including SIGIR, KDD, ICDE, WWW, ICML, AAAI, IJCAI, TKDE, and TOIS. His work on recommendation has received the Best Paper Award Nomination in ICSOC. Moreover, he has served as PC member of several top conferences including SIGIR, WWW, KDD, IJCAI, AAAI, WSDM, etc., and the regular reviewer for journals including TKDE, TNNLS, TIFS, etc.
\end{IEEEbiography}


\begin{IEEEbiography}[{\includegraphics[width=1in,height=1.25in,clip,keepaspectratio]{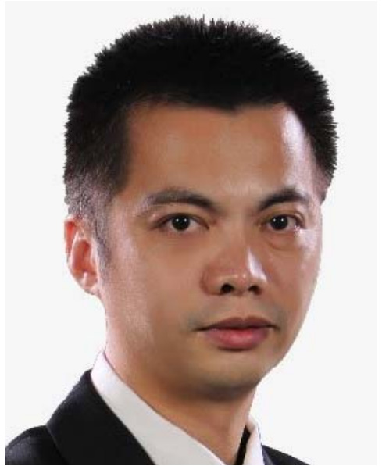}}]{Zibin Zheng} is currently a Professor and the Dean of the School of Software Engineering, at Sun Yat-sen University, Guangzhou, China. He authored or co-authored more than 200 international journal and conference papers, including one ESI hot paper and ten ESI highly cited papers. According to Google Scholar, his papers have more than 54,000 citations. His research interests include blockchain, software engineering, and services computing. He was the BlockSys’19 and CollaborateCom16 General Co-Chair, SC2’19, ICIOT18 and IoV14 PC Co-Chair. He is a Fellow of the IEEE and the IET. He was the recipient of several awards, including the Top 50 Influential Papers in Blockchain of 2018, the ACM SIGSOFT Distinguished Paper Award at ICSE2010, the Best Student Paper Award at ICWS2010.
\end{IEEEbiography}


\end{document}